\documentclass[10pt,journal,compsoc]{IEEEtran}
\usepackage[pagebackref=true,breaklinks=true,letterpaper=true,colorlinks,bookmarks=false]{hyperref}
\usepackage{times}
\usepackage{epsfig}
\usepackage{graphicx}
\usepackage{amsmath}
\usepackage{amssymb}
\usepackage{breqn}
\usepackage[bottom]{footmisc}

\usepackage[]{algorithm}
\usepackage[noend]{algpseudocode}

\makeatletter
\def\BState{\State\hskip-\ALG@thistlm}
\makeatother
\newcommand{\Pexp}{\mathbin{\text{$\vcenter{\hbox{\textcircled{$+$}}}$}}}
\newcommand{\sym}{\mathbin{\text{$\vcenter{\hbox{\textcircled{$*$}}}$}}}
\newcommand{\eadd}{\mathbin{\text{$\vcenter{\hbox{\textcircled{$+$}}}$}}}

\usepackage{pifont}
\newcommand{\cmark}{\text{\ding{51}}}
\newcommand{\xmark}{\text{\ding{55}}}

\usepackage[bottom]{footmisc}
\usepackage{latexsym}
\usepackage{url}

\ifCLASSOPTIONcompsoc
  \usepackage[nocompress]{cite}
\else
  \usepackage{cite}
\fi
\ifCLASSINFOpdf
\else
\fi
\begin{document}
%
\title{Deep Exemplar Networks for VQA and VQG}
%
%
%
%

\author{Badri N. Patro,
        and~Vinay ~P. ~Namboodiri,~\IEEEmembership{Member,IEEE}
\IEEEcompsocitemizethanks{\IEEEcompsocthanksitem Badri N. Patro is with the Department
of Electrical Engineering, Indian Institute of Technology, Kanpur, Uttar Pradesh India, 208016.\protect\\
E-mail: patrobadri.iitb@gmail.com, Home page: https://badripatro.github.io/
\IEEEcompsocthanksitem Vinay P. Namboodiri is with Department
of Computer Science and Engineering, Indian Institute of Technology, Kanpur, Uttar Pradesh, India, 208016.\protect\\ E-mail: vinay.namboodiri@gmail.com, Homepage: https://www.cse.iitk.ac.in/users/vinaypn/}
}

\IEEEtitleabstractindextext{%
\begin{abstract}
In this paper we consider the problem of solving semantic tasks such as `Visual Question Answering' (VQA), where one aims to answers related to an image and `Visual Question Generation' (VQG), where one aims to generate a natural question pertaining to an image. Solutions for VQA and VQG tasks have been proposed using variants of encoder-decoder deep learning based frameworks that have shown impressive performance. Humans however often show generalization by relying on exemplar based approaches. For instance, the work by Tversky and Kahneman \cite{tversky_CP1973availability} suggests that humans use exemplars when making categorizations and decisions. In this work we propose incorporation of exemplar based approaches towards solving these problems. Specifically, we incorporate exemplar based approaches and show that an exemplar based module can be incorporated in almost any of the deep learning architectures proposed in literature and the addition of such a block results in improved performance for solving these tasks. Thus, just as incorporation of attention is now considered de facto useful for solving these tasks, similarly, incorporating exemplars also can be considered to improve any proposed architecture for solving this task. We provide extensive empirical analysis for the same through various architectures, ablations and state of the art comparisons.

\end{abstract}

}

\maketitle

\IEEEdisplaynontitleabstractindextext

%
\IEEEpeerreviewmaketitle

\IEEEraisesectionheading{\section{Introduction}\label{sec:introduction}}

The ability to answer a question pertaining to an image (such as: what animal is in the picture?) is a simple human ability that children are taught from childhood. In order to obtain AI systems with human like abilities, it is therefore important that we propose methods that achieve significant accuracy for answering varied set of questions pertaining to a wide variety of images. A similar ability that could perhaps be even more challenging is that of asking `natural' questions pertaining to an image. Given such a semantic task, the approach by the community has been to devise datasets for solving the same and consider a number of deep learning based techniques towards solving these tasks. Indeed significant progress has been attained towards solving this task as can be observed for instance by the improving performance in successive workshops for the VQA task. In contrast to the various approaches that have been considered so far, we propose in this paper that adopting an exemplar based approach as a module for solving these problems can significantly improve performance. 

There have been a number of works in cognitive science that suggest that humans use exemplars when making categorizations and decisions including work by Tversky and Kahneman ~\cite{tversky_CP1973availability}. In this model, individuals compare new stimuli with the instances already stored in memory ~\cite{Jakel_2008}~\cite{Shepard_Science1987} and obtain answers based on these exemplars.In this paper we show that based on these ideas, one can consider incorporating exemplars for solving semantic tasks like VQA and VQG. Moreover, though the nature of the VQA and VQG tasks differs considerably, we are able to show improved performance based on incorporation of exemplar module for both these tasks. 

In this paper we consider various ways for incorporating the exemplar module. This can be obtained either through using exemplars for obtaining improved embedding using supporting and opposing exemplars. Else, it can be obtained by using supporting and opposing exemplars for improving the attention. Further, we show that it could also be perhaps considered for incorporating the exemplar based feature as shown for the VQA task. This work extends our work proposed in Patro {\it et al.} \cite{Patro_CVPR2018} where we proposed initial ways to incorporate exemplar module for VQA and Patro {\it et al.} \cite{Patro_EMNLP2018MDN} that showed ways to incorporate exemplar module for VQG task. The main contribution of this work is to propose a unified exemplar module based architecture variants for both these tasks and to compare these against the other works that have so far not considered use of such an exemplar module. 

Our results show that we obtain consistently improved performance for solving these tasks and the proposed exemplar based techniques could also yield improved attention for solving these tasks that correlate well with human attention. An important aspect of our work is to show that instead of just using a supporting exemplar, if one also uses an opposing exemplar then the results show more improvement. This is indeed well motivated by the work of Frome {\it et al.} \cite{frome_ICCV2007learning} where the authors show that instead of saying that something is closer, saying 'A' is closer to 'B' as compared to 'C' is more meaningful. These ideas have been easily incorporated in our work by using a triplet loss. Further, the number of exemplars is also a crucial choice. We see that increasing exemplars till 4 yields improvements and beyond that the improvement is not significant.  

In the rest of the paper, we show specific instances of incorporating the exemplar module for the VQA and VQG task and analyse them thoroughly with detailed ablations and comparisons.

\begin{table*}[ht]
	\centering
	\caption{Overview of Exemplar based Deep Networks for VQA \& VQG methods and their various properties.}
	\vspace{-0.7em}
		\begin{tabular}{l c c c c l } \hline
			\textbf{Methods} & \textbf{Base Model} & \textbf{Attention} & \textbf{Exemplar}& \textbf{Task} & \textbf{Dataset}\\ \hline
			Neural-IQA \cite{Malinowski_NIPS2014}& generative & {\textcolor{red}{\xmark} }& { \textcolor{red}{\xmark}} &VQA & {DAQUAR,SynthQA,HumanQA}\\
			VQA \cite{VQA}& discriminative & {\textcolor{red}{\xmark} }& { \textcolor{red}{\xmark}}&VQA &  {VQAv1}\\
			DPPnet \cite{Noh_CVPR2016}   & discriminative & {\textcolor{red}{\xmark} }& {\textcolor{red}{\xmark}}&VQA & {DAQUAR,COCO-QA, VQAv1} \\
			SAN \cite{Yang_CVPR2016}     & discriminative & {\textcolor{green}{\cmark} }& { \textcolor{red}{\xmark}}&VQA & {DAQUAR,COCO-QA, VQAv1}\\
			mQA \cite{Gao_NIPS2015} & generative & {\textcolor{red}{\xmark} }& { \textcolor{red}{\xmark}}&VQA & {FM-IQA}\\
			BAYESIAN\cite{kafle_CVPR2016answer} & discriminative & {\textcolor{green}{\cmark} }& { \textcolor{red}{\xmark}}&VQA &{DAQUAR,COCO-QA, VQAv1,Visual7w}\\
			DMN \cite{Xiong_arXiv2016} & discriminative & {\textcolor{green}{\cmark} }& { \textcolor{red}{\xmark}} &VQA &{DAQUAR, bAbI-10k,VQAv1} \\ 
			QRU \cite{Li_NIPS2016} & discriminative & {\textcolor{green}{\cmark} }& { \textcolor{red}{\xmark}}&VQA &{COCO-QA, VQAv1}\\			
			HieCoAtt \cite{Lu_NIPS2016} & discriminative & {\textcolor{green}{\cmark} }& { \textcolor{red}{\xmark}}&VQA &{COCO-QA, VQAv1}\\
			ASK\cite{wu_CVPR2016ask}& generative & {\textcolor{red}{\xmark} }& { \textcolor{red}{\xmark}} &VQA &{COCO-QA, VQAv1}\\
			MCB-att\cite{Fukui_arXiv2016} & discriminative & {\textcolor{green}{\cmark} }& { \textcolor{red}{\xmark}} &VQA & {VQAv1}\\
			MLB \cite{Kim_ICLR2017} & discriminative & {\textcolor{green}{\cmark} }& { \textcolor{red}{\xmark}}&VQA  & {VQAv1}\\ 
			Natural \cite{mostafazadeh2016generating} & generative & {\textcolor{red}{\xmark}}& {\textcolor{red}{\xmark}} &VQG & {VQG (COCO, Bing, Flickr)}\\
            Creative \cite{jain2017creativity}& generative & {\textcolor{red}{\xmark} }& {\textcolor{red}{\xmark}}&VQG  & {VQG,VQAv1}\\\hline
            DAN (Our-:1+3+4+5+6+8) & discriminative & {\textcolor{green}{\cmark} }& {\textcolor{green}{\cmark}}&VQA  & {VQAv1,VQAv2}\\ 
            DCN (Our-:1+3+4+5+6+8) & discriminative & {\textcolor{green}{\cmark} }& { \textcolor{green}{\cmark}}&VQA & {VQAv1,VQAv2}\\ 
            DJN (Our-:1+3+4+5+7+8) & discriminative & {\textcolor{red}{\xmark} }& { \textcolor{green}{\cmark}}&VQA &  {VQAv1,VQAv2}\\
            MDN-att (Our-:1+2+4+5+6+9)  & generative & {\textcolor{green}{\cmark} }& {\textcolor{green}{\cmark}}&VQG &  {VQG,VQAv1}\\
            MDN-joint (Our-:1+2+4+5+7+9) & generative & {\textcolor{red}{\xmark} }& {\textcolor{green}{\cmark}}&VQG & {VQG,VQAv1}\\ \hline
		\end{tabular}

	\vspace{-1em}
	\label{tab:overview}
\end{table*}

Through this paper we provide the following contributions
\begin{itemize}
    \item We propose an exemplar based approach to improve visual question answering (VQA) and visual question generation (VQG) methods.
    \item We evaluate various modifications of incorporating exemplars through attention and context. The overview of the same is provided in figure-1. In Table-1 indicates the variants that we propose.
    \item We show that these methods correlate better with human attention and result in an improved VQA and VQG system that improves over the state-of-the-art for image based attention methods. It is also competitive with respect to other proposed methods for this problem.
\end{itemize}

\begin{figure}[ht]
	\centering
	\includegraphics[width=0.49\textwidth]{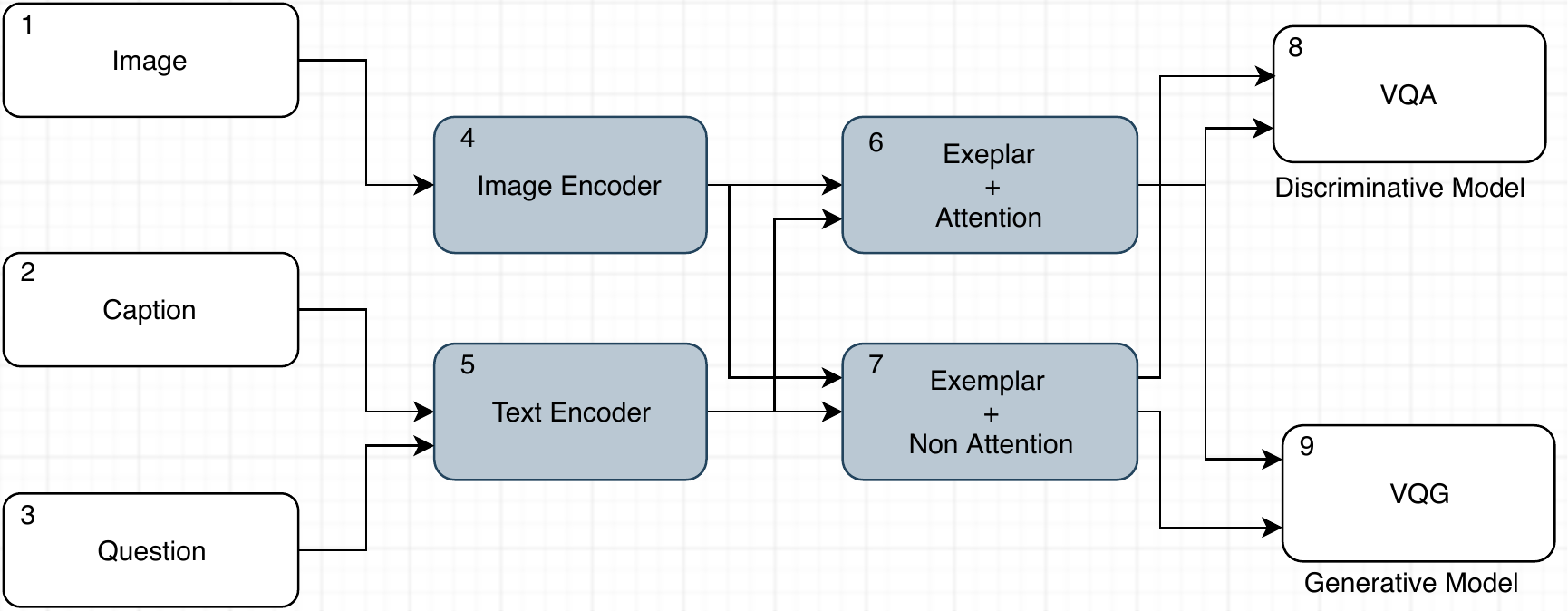}
	\vspace{-1em}
	\caption{ This is an overview of our Deep Exemplar Network for VQA and VQG. The numbers in the boxes are used to specify the architecture compared in table 1. }
	\vspace{-0.7em}
	\label{fig:DEN}
\end{figure}



\section{Related Work}
\label{sec:lit_surv}
There has been extensive work done in the Vision and Language domain for solving image captioning, paragraph generation, Visual Question Generation (VQG) and Visual Dialog.~\cite{Barnard_JMLR2003,Farhadi_ECCV2010,Kulkarni_CVPR2011} proposed conventional machine learning methods for image description.~\cite{Socher_TACL2014,Vinyals_CVPR2015,Karpathy_CVPR2015,Xu_ICML2015,Fang_CVPR2015,Chen_CVPR2015,Johnson_CVPR2016,Yan_ECCV2016} have generated descriptive sentences from images with the help of Deep Networks. There have been many works for solving Visual Dialog~\cite{chappell_HFES2004,Das_EMNLP2016,visdial,de2017guesswhat,strub2017end}. 
However, the problem of Visual Question Answering (VQA) is a recent problem that was initiated as a new kind of visual Turing test. The aim was to show the progress of systems in solving even more challenging tasks as compared to the traditional visual recognition tasks such as object detection and segmentation. Initial work in this area was by Geman {\it et al.} \cite{Geman_PNAS2015} that proposed this visual Turing test. Around the same time, Malinowski {\it et al.} \cite{Malinowski_NIPS2014} proposed a multi-world based approach to obtain questions and answer them from images. These works aimed at answering questions of a limited type. In this work, we aim at answering free-form open-domain\cite{VQA} questions as was attempted by later works.

An initial approach towards solving this problem in the open-domain form was by  \cite{Malinowski_ICCV2015}. This was inspired by the work on neural machine translation that proposed translation as a sequence to sequence encoder-decoder framework \cite{Sutskever_NIPS2014}. However, subsequent works \cite{Ren_NIPS2015}\cite{VQA} approached the problem as a classification problem using encoded embeddings. They used soft-max classification over an image embedding  (obtained by a CNN) and a question embedding (obtained using an LSTM). Further work by Ma {\it et al.} \cite{Ma_AAAI2016} varied the way to obtain an embedding by using CNNs to obtain both image and question embeddings. Another interesting approach \cite{Noh_CVPR2016} used dynamic parameter prediction where weights of the CNN model for the image embedding are modified based on the question embedding using hashing.  These methods however, are not attention based. The use of attention enables us to focus on specific parts of an image or question that are pertinent, for instance and also offer valuable insight into the performance of the system.

There has been significant interest in including attention to solve the VQA problem. Attention based models comprise of image based attention models, question based attention, and some that are both image and question based attention. In image based attention approach the aim is to use the question in order to focus attention over specific regions in an image \cite{Shih_CVPR2016}. An interesting recent work \cite{Yang_CVPR2016} has shown that it is possible to repeatedly obtain attention by using stacked attention over an image based on the question. Our work is closely related to this work. There have been further work \cite{Li_NIPS2016} that considers a region based attention model over images. The image based attention has allowed systematic comparison of various methods as well as enabled analysis of the correlation with human attention models, as shown by \cite{Das_EMNLP2016}.  In our approach, we focus on image based attention using differential attention and show that it correlates better with image based attention. There have been a number of interesting works on question based attention as well as \cite{Zhu_CVPR2016}\cite{Xu_ECCV2016}. An exciting work obtains a varied set of modules for answering questions of different types of \cite{andreas16naacl}. Recent work also explores joint image and question based hierarchical co-attention \cite{Lu_NIPS2016}. The idea of differential attention can also be explored through these approaches.  However, we restrict ourselves to image based attention as our aim is to obtain a method that correlates well with human attention~\cite{Das_EMNLP2016}. There has been an interesting work by \cite{Fukui_arXiv2016} that advocates multimodal pooling and obtains state of the art in VQA. Interestingly, we show that by combining it with the proposed method further improves our results.

Generating a natural and engaging question is an interesting and challenging task for a smart robot (like chat-bot). It is a step towards having a natural visual dialog instead of the widely prevalent visual question answering bots. Further, having the ability to ask natural questions based on different contexts is also useful for artificial agents that can interact with visually impaired people. While the task of generating questions automatically is well studied in the NLP community, it has been relatively less studied for image-related natural questions. This is still a difficult task~\cite{mostafazadeh2016generating} that has gained recent interest in the community.

Recently there have been many deep learning based approaches as well for solving the text-based question generation task such as as~\cite{Du_ArXiv2017}.
Further,~\cite{Serban_Arxiv2016} has proposed a method to generate a factoid based question based on triplet set \{subject, relation, and object\} to capture the structural representation of text and the corresponding generated question.

However, Visual Question Generation (VQG) is a separate task that is of interest in its own right and has not been so well explored~\cite{mostafazadeh2016generating}. This is a vision based novel task aimed at generating natural and engaging question for an image.
~\cite{Yang_arXiv2015} proposed a method for continuously generating questions from an image and subsequently answering those questions.
The works closely related to ours are that of~\cite{mostafazadeh2016generating} and~\cite{jain2017creativity}. In the former work, the authors used an encoder-decoder based framework whereas in the latter work,
The authors extend it by using a variational autoencoder based sequential routine to obtain natural questions by performing sampling of the latent variable. The methods proposed so far have not considered an exemplar based approach that we analyse in this work.
\begin{figure}[ht]
	\centering
	\includegraphics[width=0.45\textwidth]{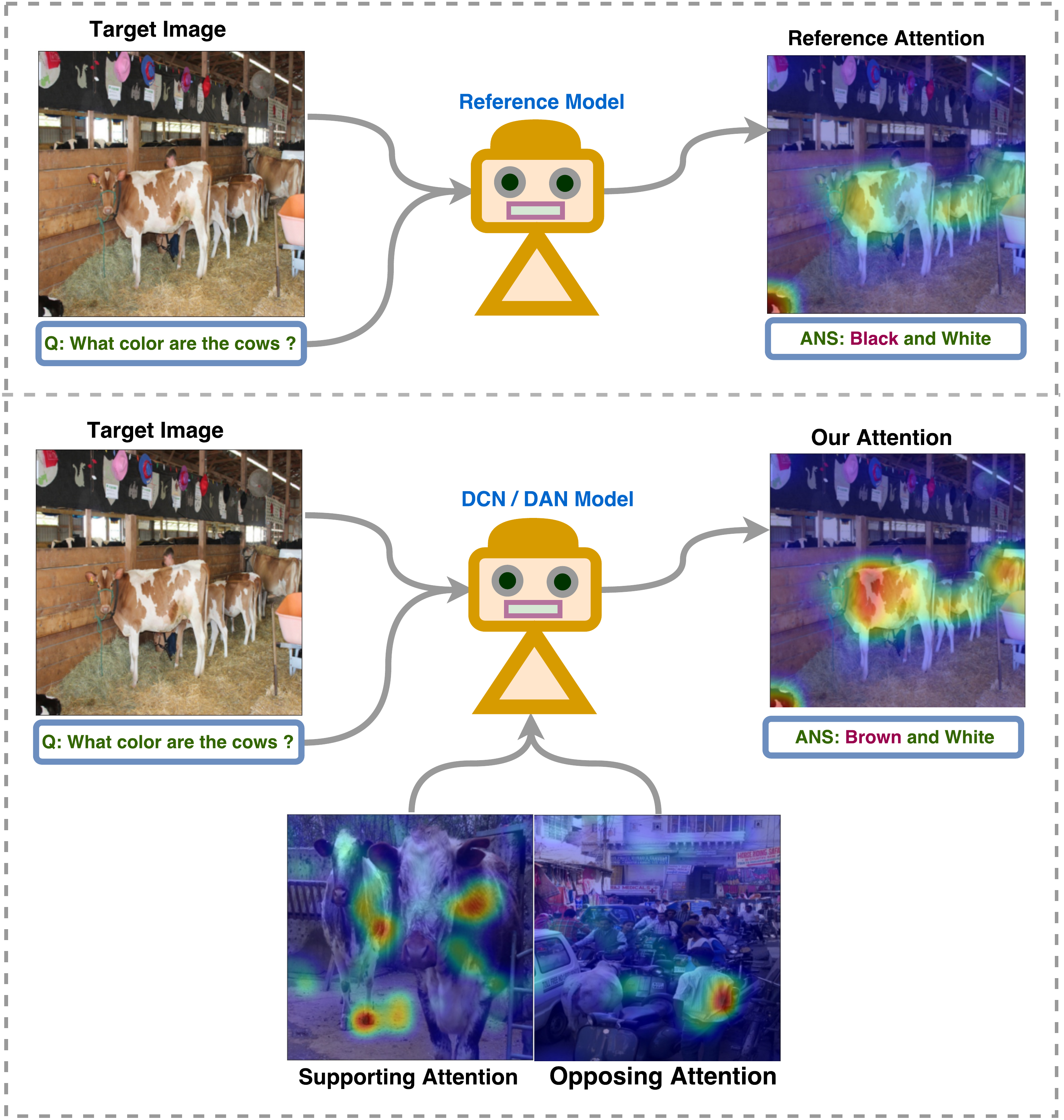}
	\caption{ Illustration of improved attention obtained using Differential Context Network. Using the baseline reference we got answer as: ``Black and White'' But, using our methods DAN or DCN we get answer as ``Brown and White'', that is actually the color of the cow. We provide the attention map that indicates the actual improvement in attention.}
	\label{fig:flow}
\end{figure}

 \begin{figure*}[ht]
     \small
     \centering
     \begin{tabular}[b]{ c  c  }
     (a) DAN for VQA  & (b) MDN-attention for VQG  \\ 
     \includegraphics[width=0.5\textwidth,height=0.35\textwidth]{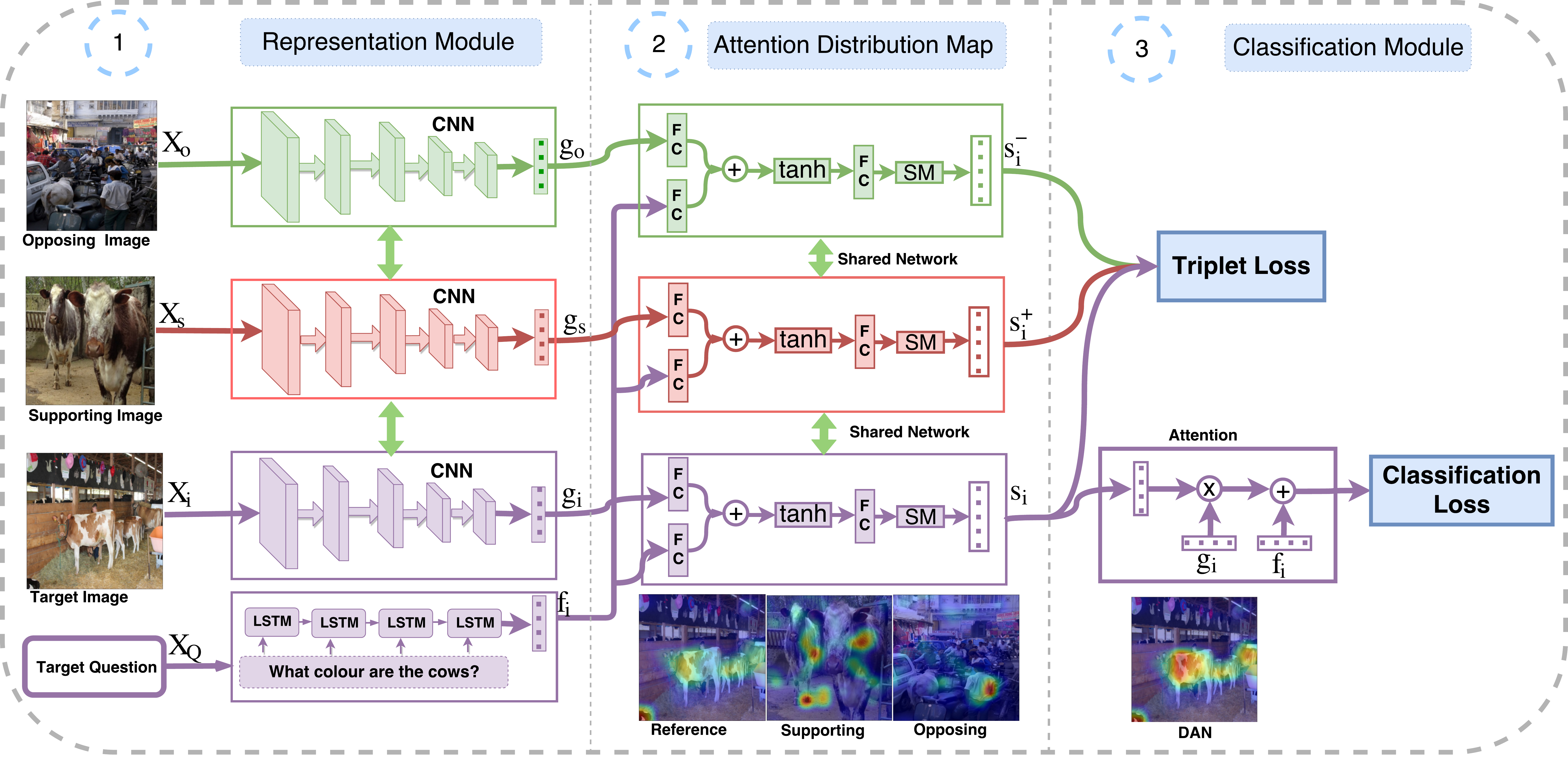}
     & \includegraphics[width=0.5\textwidth,height=0.35\textwidth]{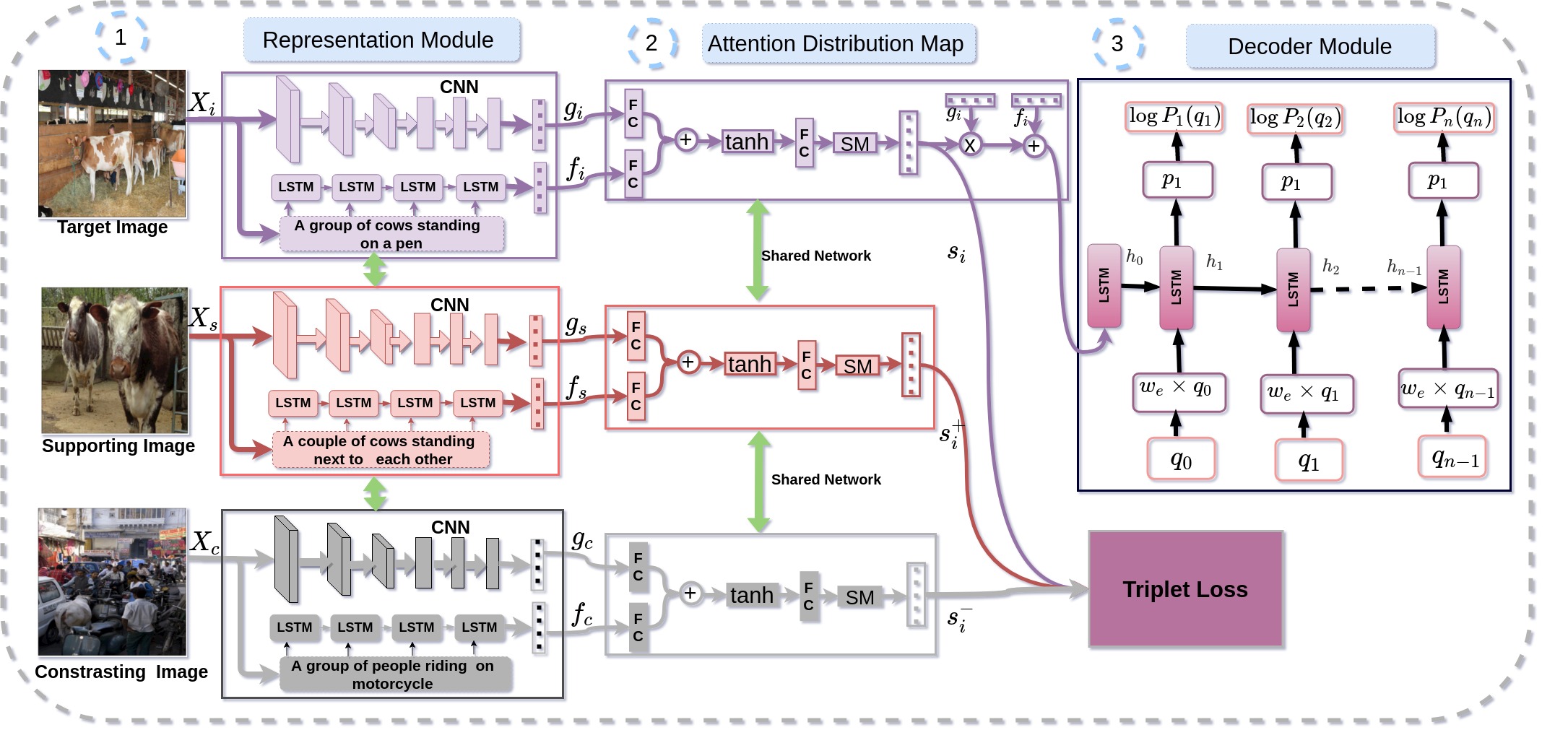}
     \end{tabular}
       \vspace{-1em}
      \caption{This is an overview of our Attentive Exemplar for VQA (Differential Attention Network (DAN)) and for VQG (Multimodal Differential Network (MDN-attention) ). It consists of 3 parts: 1) Representation Module which provides embedding representation of the target, supporting, \& opposing image  using CNN and also encodes a given question sentence using LSTM, 2) Attention Distribution Map which combines each image embedding with question encoding feature using attention mechanism and 3) Classification Module which classifies the differential attention feature into one of the answer classes for VQA and Decoder module for VQG. Finally, this whole network is trained with triplet and cross entropy loss. } 
      \label{fig:DAN}
        \vspace{-1.5em}
 \end{figure*}
\section{Method}\label{sec:method}
We provide two types of exemplar method to solve VQA and VQG task. The first one is an attentive exemplar, and the second one is the fused exemplar method. We explain both the methods in the following sub-sections. A common requirement for both these tasks is to obtain the nearest semantic exemplars. 

\subsection{Finding Exemplars}
In our method, we use semantic nearest neighbors. Image level similarity does not suffice as the nearest neighbor may be visually similar but may not have the same context implied in the question (for instance, `Are the children playing?' produces similar results for images with children based on visual similarity, whether the children are playing or not). In order to obtain semantic features, we use a VQA system \cite{Lu2015} to provide us with a joint image-question level embedding that relates meaningful exemplars. We compared image-level features against the semantic nearest neighbors and observed that the semantic nearest neighbors were better. We used the semantic nearest neighbors in a k-nearest neighbor approach using a K-D tree data structure to represent the features. The ordering of the data-set features is based on the Euclidean distance. In section~\ref{sec:param_analysis} we provide the evaluation with several values of nearest neighbors that were used as supporting exemplars. For obtaining opposing exemplar, we used a far neighbor that was an order of magnitude further than the nearest neighbor. This we obtained through a coarse quantization of training data into bins. We specified the opposing exemplar as one that was around 20 clusters away in a 50 cluster ordering. This parameter is not stringent, and it only matters that the opposing exemplar is far from the supporting exemplar. We show that using these supporting and opposing exemplars aids the method, and any random ordering adversely effects the method.

\subsection{Attentive Exemplar Approach}\label{subsec:method}
In this paper, we adopt a classification framework for VQA that uses the image embedding combined with the question embedding to solve for the answer using a softmax function in a multiple-choice setting. A similar setting is adopted in the Stacked Attention Network (SAN) \cite{Yang_CVPR2016}, which also aims at obtaining better attention and several other state-of-the-art methods. We provide two different variants for obtaining differential attention in the VQA system. We term the first variant a `Differential Attention Network' (DAN) and the other a `Differential Context Network' (DCN), which improves on the DAN method. Also, we extend the DAN method for the generative framework (VQG). We name that MDN-attention Method. 
Our (MDN-attention) method is based on a sequence to sequence network ~\cite{Sutskever_NIPS2014,Vinyals_CVPR2015,Bahdanau_arXiv2014}. The sequence to sequence network has a text sequence as input and output. In our method, we take an image as input and generate a natural question as output. The architecture for our model is shown in Figure~\ref{fig:DAN}(b) and Figure~\ref{fig:MDN}(b). Our model contains three main modules, (a) Representation Module that extracts multimodal features (b) Mixture Module that fuses the multimodal representation and (c) Decoder that generates question using an LSTM-based language model.

\subsubsection{Attentive Exemplar for VQA : Differential Attention Network (DAN)}
In the DAN method, we use a multi-task setting. As one of the tasks, we use a triplet loss\cite{Hoffer_Springer2015} to learn a distance metric. This metric ensures that the distance between the attention weighted regions of near examples is less, and the distance between attention weighted far examples are more. The other task is the main task of VQA.
More formally, given an image $x_i$, we obtain an embedding $g_i$ using a CNN that we parameterized through a function $G(x_i, W_c)$ where $W_c$ are the weights of the CNN. Similarly, the question $q_i$ results in a question embedding $f_i$ after passing through an LSTM parameterised using the function $F(q_i, W_l)$ where $W_l$ are the weights of the LSTM. This is illustrated in part 1 of figure-~\ref{fig:DAN}(a). The output image embedding $g_i$ and question embedding $f_i$ are used in an attention network that combines the image and question embeddings with a weighted softmax function and produce output attention weighted vector $s_i$. The attention mechanism is illustrated in figure-~\ref{fig:DAN}(a). The weights of this network are learned end-to-end learning using the two losses, a triplet loss and a soft-max classification loss for the answer (shown in part 3 of figure-~\ref{fig:DAN}(a)).  The aim is to obtain attention weight vectors that bring the supporting exemplar attention close to the image attention and far from the opposing exemplar attention. The joint loss function used for training is given by:


\begin{equation}
\begin{split}
\label{eq1}
&    L( \textbf{s},\textbf{y},\theta) = \frac{1}{N}\sum^{N}_{i=1}\left( L_{cross}( \textbf{s},\textbf{y}) + \nu T(s_i,s_i^+, s_i^-) \right)\\
& L_{cross}( \textbf{s},\textbf{y}) = -\frac{1}{C}\sum_{j=1}^{C} y_{j} \texttt{log} \texttt{p}(c_{j}|\textbf{s}) 
    \end{split}
\end{equation}
Where $\theta$ is the set of model parameters for the two loss functions, $y$ is the output class label and $s$ is the input sample. $C$ is the total number of classes in VQA ( consists of the set of total number of output classes including color, count etc. ) and  $N$ is the total number of samples. The first term is the classification loss and the second term is the triplet loss. $\nu$ is a constant that controls the ratio between classification loss and triplet loss.  $T(s_i, s_i^{+}, s_i^{-})$ is the triplet loss function that is used. This is decomposed into two terms, one that brings the positive sample closer and one that pushes the negative sample farther. This is given by
\begin{dmath}
\label{eq2}
T(s_i,s_i^+,s_i^-) = \texttt{max}(0, ||t(s_{i})-t(s_{i}^{+})||^{2}_{2} + \alpha - ||t(s_{i})-t(s_{i}^{-})||^{2}_{2})
\end{dmath}
The constant $\alpha$ controls the separation margin between supporting and opposing exemplars. The constants $\nu$ and $\alpha$ are obtained through validation data.

The method is illustrated in figure-~\ref{fig:DAN}. We further extend the model to a quintuplet setting where we bring two supporting attention weights closer and two opposing attention weights further in a metric learning setting. We observe in section~\ref{sec:param_analysis} that this further improves the performance of the DAN method.

\begin{figure}[ht]
	\centering
	\includegraphics[width=0.49\textwidth]{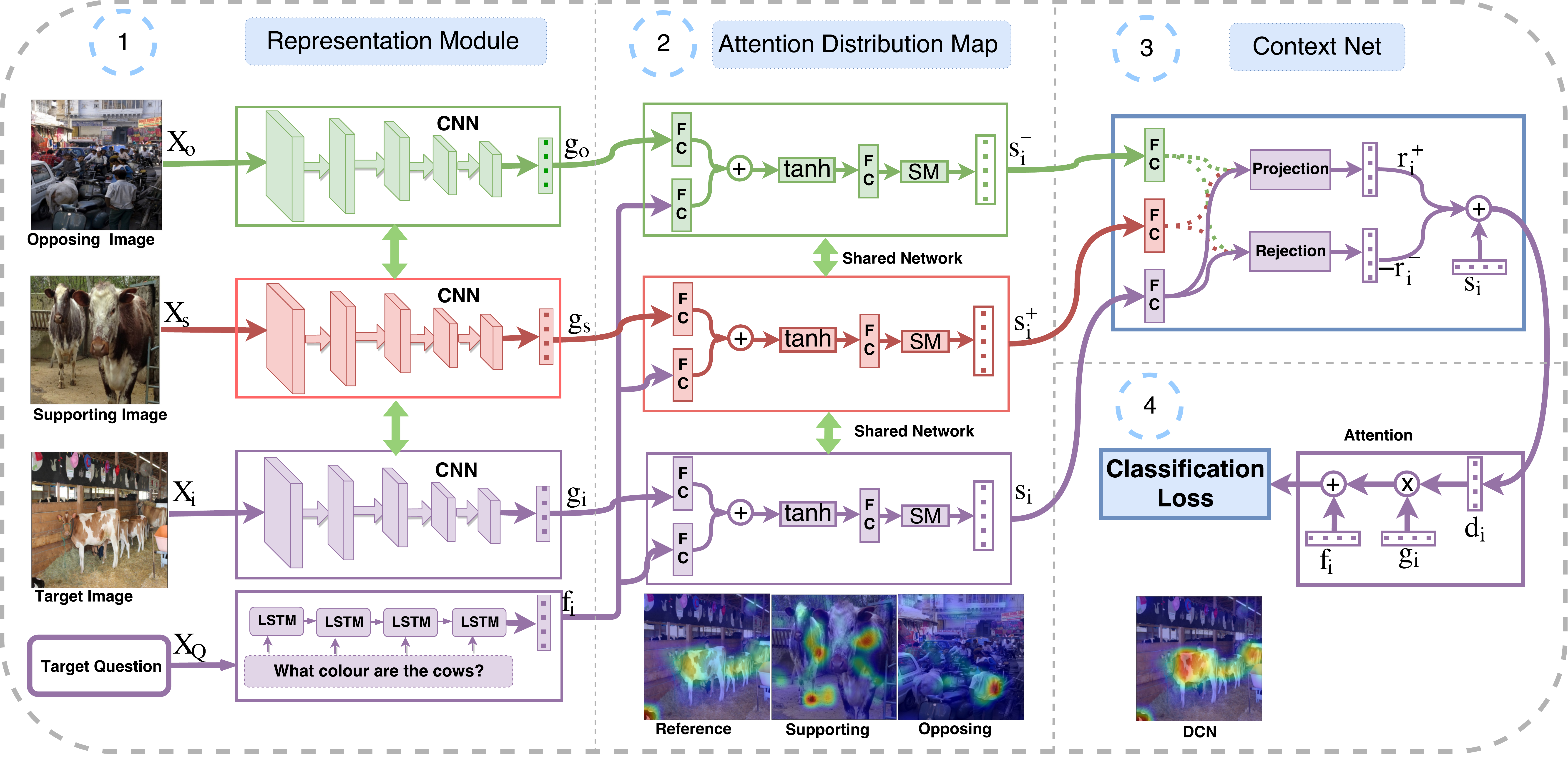}
	\vspace{-1.1em}
	\caption{ This is an overview of our Differential Context Network. It consists of 4 parts: 1) Representation Module which provides an embedding representation of the target, supporting, and opposing image  using CNN and also encodes a given question sentence using LSTM, 2) Attention Distribution Map which combines each image embedding with question encoding feature using attention mechanism, 3) Context Net which provides projection and rejection context of the attention feature and 4) Classification Module which classifies the final context feature into one of the answer classes. Finally, this whole network is trained with triplet and cross entropy loss. }
	\label{fig:DCN}
	\vspace{-0.7em}
\end{figure}

\begin{figure*}[ht]
     \small
     \centering
     \begin{tabular}[b]{ c  c  }
     (a) DJN for VQA  & (b) DJN for VQG  \\ 
     \includegraphics[width=0.45\textwidth]{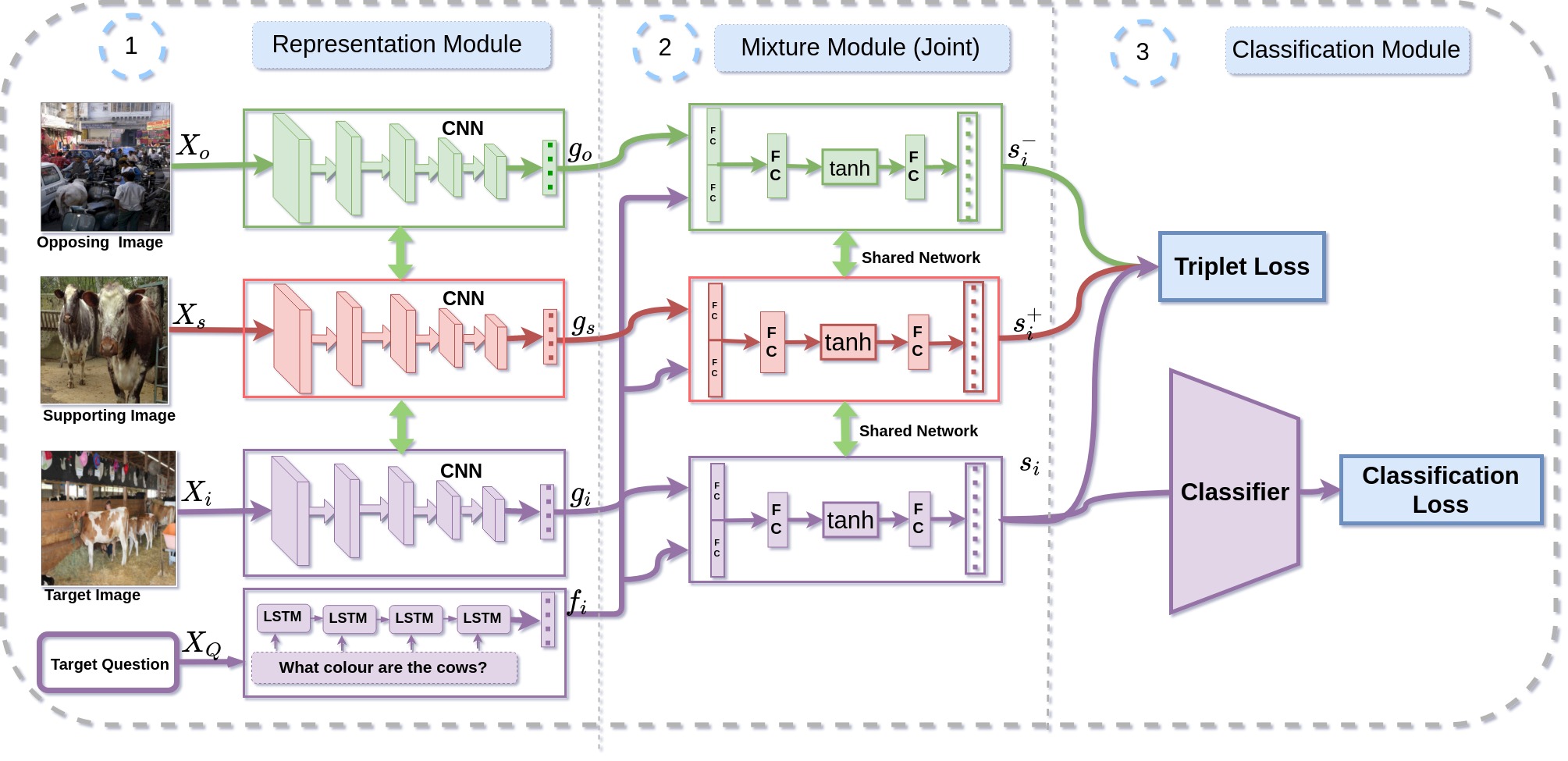}
     & \includegraphics[width=0.5\textwidth]{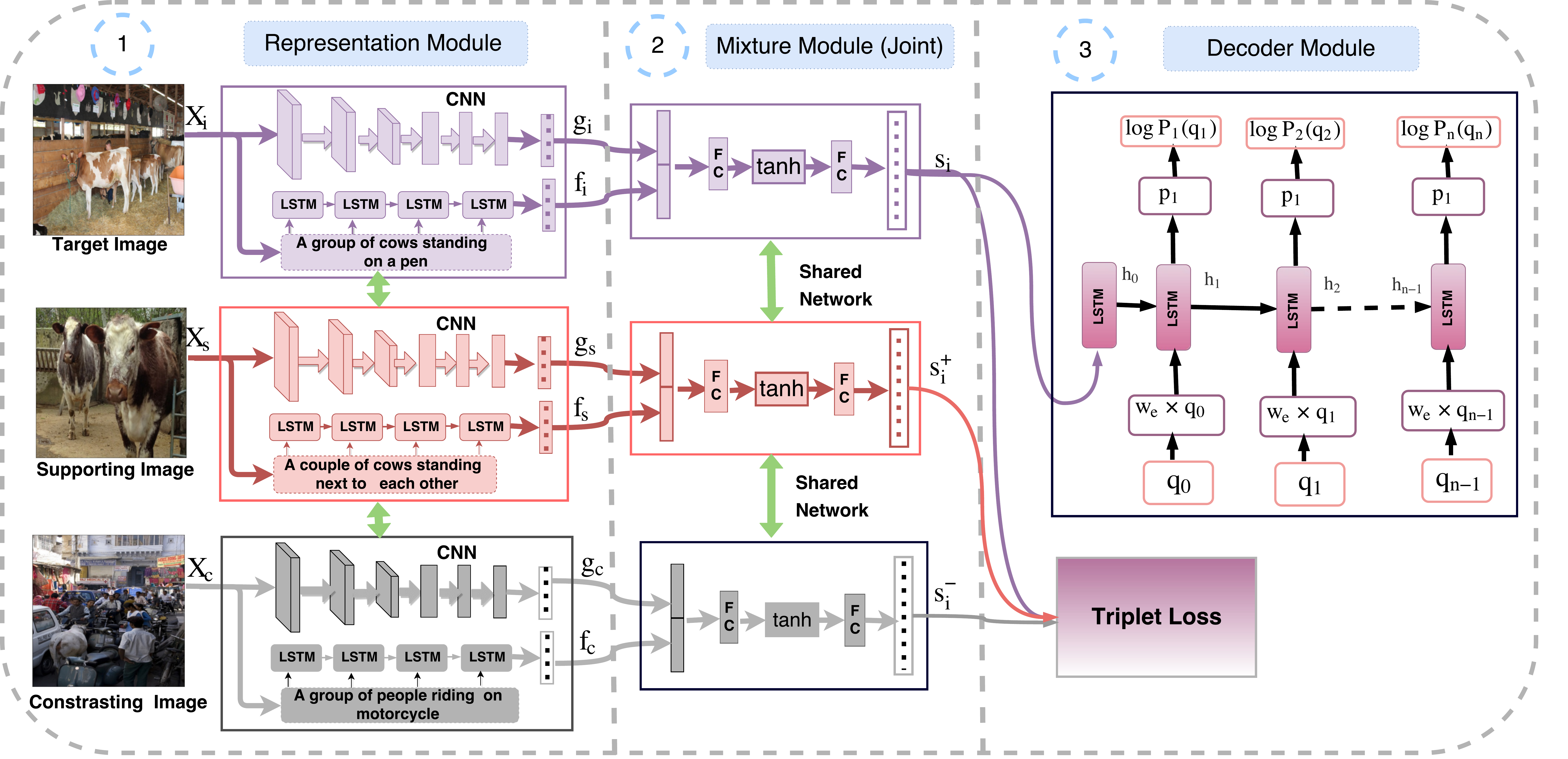}
     \end{tabular}
     \vspace{-1.2em}
      \caption{This is an overview of our Multimodal Differential Network for Visual Question Generation. It consists of a Representation Module which extracts multimodal features, a Mixture Module that fuses the multimodal representation and a Decoder that generates question using an LSTM based language model. In this figure, we have shown the Joint Mixture Module. We train our network with a Cross-Entropy and Triplet Loss. } 
      \label{fig:MDN}
        \vspace{-1.5em}
 \end{figure*}

\textbf{Differential Context Network (DCN)}
We next, consider the other variant that we propose where the differential context feature is added instead of only using it for obtaining attention. The first two parts are the same as those for the DAN network. In part 1, we use the image, the supporting, and the opposing exemplar and obtain the corresponding image and question embedding. This is followed by obtaining attention vectors $s_i, s_i^+, s_i^-$ for the image, the supporting, and the opposing exemplar. While in DAN, these were trained using a triplet loss function, in DCN, we obtain two context features, the supporting context $r_i^+$ and the opposing context $r_i^-$. This is shown in part 3 in figure-~\ref{fig:DCN}. The supporting context is obtained using the following equation,

\begin{equation}
\label{eq3}
r_i^+ = (s_i \bullet s_i^+)\frac{s_i} {\lVert s_i \rVert_{L_{2}}^{2}} + (s_i \bullet s_i^-)\frac{s_i} {\lVert s_i \rVert_{L_{2}}^{2}}
\end{equation}




Where $\bullet$ is the dot product. This results in obtaining correlations between the attention vectors. 
The first term of the supporting context $r_i^+$ is the vector projection of $s_i^+$ on  $s_i$ and and second term is the vector projection of $s_i^-$ on  $s_i$. Similarly, for opposing context we compute  vector projection of $s_i^+$ on  $s_i$ and $s_i^-$ on  $s_i$.  The idea is that the projection measures similarity between the vectors that are related. We subtract the vectors that are not related from the resultant. While doing so, we ensure that we enhance similarity and only remove the feature vector that is not similar to the original semantic embedding. This equation provides the additional feature that is supporting and is relevant for answering the current question $q_i$ for the image $x_i$.

Similarly, the opposing context is obtained by the following equation

\begin{equation}
\label{eq4}
r_i^- = (s_i^+ - (s_i \bullet s_i^+)\frac{s_i} {\lVert s_i \rVert_{{2}}^{2}}) + (s_i^- - (s_i \bullet s_i^-)\frac{s_i} {\lVert s_i \rVert_{{2}}^{2}})
\end{equation}

We next compute the difference between the supporting and opposing context features i.e. $r_i^+ - r_i^-$ that provides us with a differential context feature $\hat{d_i}$. This is then either added with the original attention vector $s_i$ (DCN-Add) or multiplied (DCN-Mul) providing us with the final differential context attention vector $d_i$. This is then the final attention weight vector multiplied to the image embedding $g_i$ to obtain the vector $v_i$ that is then used with the classification loss function. This is shown in part 4 in the figure-~\ref{fig:DCN}. The resultant attention is observed to be better than the earlier differential attention feature obtained through DAN as the features are also used as context. The network is trained end-to-end using the following soft-max classification loss function.
\begin{equation}
\label{eq5}
    L( \textbf{v},\textbf{y},\theta) = -\sum_{j=1}^{C} y_{j} \texttt{log} \texttt{p}(c_{j}|\textbf{v}) 
\end{equation}

\subsubsection{Attentive Exemplar for VQG : Multimodal Differential Network (MDN-attention)}\label{subsec:mdn_attention}
In our method, we take an image as input and generate a natural question as output. The architecture for our model is shown in Figure~\ref{fig:DAN}(b). The proposed Multimodal Differential Network (MDN-attention) consists of a representation module, an attention distribution module Module, and a decoder module to generate a question sentence. 


\textbf{Representation Module}
We use a triplet network~\cite{Frome_ICCV2007,Hoffer_Springer2015} in our representation module. 
We referred to a similar kind of work done in \cite{Patro_CVPR2018} for building our triplet network. 
The triplet network consists of three sub-parts: target, supporting, and contrasting networks. All three networks share the same parameters. Given an image $x_i$, we obtain an embedding $g_i$ using a CNN parameterized by a function $G(x_i, W_c)$ where $W_c$ are the weights for the CNN. The caption $C_i$ results in a caption embedding $f_i$ through an LSTM parameterized by a function $F(C_i, W_l)$ where $W_l$ are the weights for the LSTM. This is shown in part 1 of Figure~\ref{fig:MDN}. 
Similarly we obtain image embeddings $g_s$ \& $g_c$  and  caption embeddings $f_s$ \& $f_c$.  
\begin{equation}
 \begin{split}
&g_i=G(x_i,W_c)=CNN(x_i)\\
&f_i=F(C_i, W_l)=LSTM(C_i)
\end{split}
\end{equation}

The output of the attention method $F_{Ma}$ is the weighted average of attention probability vector $a_{p}$ and convolutional features $F_{Ia}$. The attention probability vector decides the contribution of each convolutional feature based on the given context vector. The equation we use is as follows:
\begin{equation}
    \begin{split}
        & s_{i} = W^{T}_{a} *\tanh(W_{ce} * F_{Ci} + W_{ie} *  F_{Ia} +  b_a) \\
        & a_{p} = \text{softmax}(s_i) \\
        & F_{Ma} = a^{T}_{p}*F_{Ia}
    \end{split}
\end{equation}
where $F_{Ia}$ is  the 14x14 512-dimensional  convolution feature map from the fifth convolution layer of VGG-19 Net \cite{simonyan_arXiv2014} of image $x_{i}$ and $F_{Ci}$ is the context vector for same image $x_i$.The attention probability vector is a 196-dimensional unit vector. $W_{a},W_{ce},W_{ie}, b_a$ are the weights and bias of different layers.

\textbf{Mixture Module}\label{mixture_model}
The mixture module brings the image and caption embeddings to a common feature embedding space. The input to the module is the embeddings obtained from the representation module. We use an attention module similar to the attention method used in visual question answering (VQA) for the mixture module to obtain common feature embedding. The attention  receives image features $g_i$ \& the caption embedding $f_i$, and outputs a fixed dimensional feature vector $s_i$. The attention method concatenates $g_{i}$  \& $f_{i}$ and maps them to a fixed-length feature vector $s_{i}$ as follows:
\begin{equation}
    \begin{split}
        & s_{i} = W^{T}_{a} *\tanh(  W_{ia}  g_{i} + (W_{ca}  f_{i} + b_a) \\
        & \alpha = \text{softmax}(s_i) \\
        & e_{i} = \sum^{}_{}{\alpha*g_{i}} + f_{i}
    \end{split}
\end{equation}
where $g_{i}$ is  the 4096-dimensional  convolutional feature from the  FC7 layer of pretrained VGG-19 Net~\cite{simonyan_arXiv2014}. $W_{ia}, W_{ca},W_{a}$ are the weights and $b_a$ is the bias for different attention layers. Finally, the attention probability combine with image feature to obtain final encoding feature $e_{i}$. 

 Similarly, we obtain context vectors  $s^+_i$ \& $s^-_i$ for the supporting and contrasting exemplars. Details for other fusion methods are present in supplementary. The aim of the triplet network~\cite{Schroff_CVPR2015} is to obtain context vectors that bring the supporting exemplar embeddings closer to the target embedding and vice-versa.
This is obtained as follows:
\begin{equation}
 \begin{split}
& D(t(s_{i}),t(s_{i}^{+})) +\alpha < D(t(s_{i}),t(s_{i}^{-}))\\
& \forall{(t(s_{i}),t(s_{i}^{+}),t(s_{i}^{-}))} \in M,
\end{split}
\end{equation}
Where $D(t(s_{i}),t(s_{j})) = ||t(s_{i})- t(s_{j})||_{2}^{2}$ is the Euclidean distance between two embeddings $t(s_{i})$ and  $t(s_{j})$. M is the training dataset that contains all set of possible triplets. $T(s_i, s_i^{+}, s_i^{-})$ is the triplet loss function. This is decomposed into two terms, one that brings the supporting sample closer and one that pushes the contrasting sample further. This is given by, 
\begin{equation} 
    T(s_i,s_i^+,s_i^-) =\texttt{max}(0,  D^{+} + \alpha - D^{-})
    \label{triplet_loss}
\end{equation}
\noindent Here $D^+, D^-$ represents the Euclidean distance between the target and supporting sample, and target and opposing sample, respectively. The parameter $\alpha(=0.2)$ controls the separation margin between these 
and is obtained through validation data.

\textbf{Decoder: Question Generator}
\noindent The role of the decoder is to predict the probability for a question, given $s_i$. RNN provides a nice way to perform conditioning on previous state values using a fixed-length hidden vector. 
The conditional probability of a question token at a particular time step ${q_{t}}$ is modeled using an LSTM as used in machine translation~\cite{Sutskever_NIPS2014}. 
At time step $t$, the conditional probability is denoted by $P( {q_{t}} | {I ,C},{q_0},...{q_{t-1}})= P( {q_{t}} | {I ,C},h_{t})$, where $h_{t}$ is the hidden state of the LSTM cell at time step $t$, which is conditioned on all the previously generated words $\{{q_0},{q_1},...{q_{N-1}}\}$.
The word with maximum probability in the probability distribution of the LSTM cell at step $k$  is provided as an input to the LSTM cell at step $k+1$, as shown in part 3 of Figure~\ref{fig:DAN}(b) and Figure-~\ref{fig:MDN}(b). At $t=-1$, we are feeding the output of the mixture module to LSTM. $\hat{Q}=\{\hat{q_0},\hat{q_1},...\hat{q_{N-1}}\}$ are the predicted question tokens for the input image $I$. Here, we are using $\hat{q_0}$ and $\hat{q_{N-1}}$ as the special token START and STOP, respectively. 
The softmax probability for the predicted question token at different time steps is given by the following equations where LSTM refers to the standard LSTM cell equations:
\begin{equation*}
 \begin{split}
& x_{-1}=e_i=\mbox{Mixture Module}(g_{i},f_{i}) \\
& h_0=\mbox{LSTM}(x_{-1})\\
& x_t=W_e*q_t,  \forall t\in \{0,1,2,...N-1\} \\
& {h_{t+1}}=\mbox{LSTM}(x_t,h_{t}), \forall t\in \{0,1,2,...N-1\}\\
& o_{t+1} = W_o * h_{t+1} \\
& \hat{y}_{t+1} = P( q_{t+1} | {I ,C},h_{t})= \mbox{Softmax}(o_{t+1})\\
& Loss_{t+1}=loss(\hat{y}_{t+1},y_{t+1})
 \end{split}
\end{equation*}
 Where  $\hat{y}_{t+1}$ is the probability distribution over all question tokens. $loss$ is cross entropy loss.


\begin{table}[htb]
	\caption{Analysis network parameter for DJN  on VQA1.0 Open-Ended (test-dev)}
	\vspace{-1em}
	\label{tab:djn_param}
    \centering
    \begin{tabular}{l|cccr  } \hline
    	\textbf{Models} & \textbf{All} & \textbf{Yes/No} & \textbf{Number} & \textbf{others} \\ \hline 	
    	 	LSTM Q+I(LQI)  & { 52.3 }& { 78.6 }& { 35.2 }& {35.6} \\ \hline
    		DAN(K=1)+LQI  & {55.2 }& { 79.4 }& { 35.8 }& { 36.8}\\ 				
    		DAN(K=2)+LQI  & {56.9 }& { 79.9 }& { 35.9 }& { 38.2 } \\ 				
    		DAN(K=3)+LQI  & {57.5 }& { 80.1 }& { 36.0 }& \textbf{ 39.5 }\\ 
    		DAN(K=4)+LQI  & \textbf{58.1 }& \textbf{ 80.4 }& \textbf{ 36.1 }& { 40.2 } \\ \hline 
    		MCB & {60.8}& { 81.4 }& {35.4 }& { 47.2} \\ 
            DAN(K=1)+MCB  & {61.2 }& { 81.5 }& { 36.3 }& { 49.2} \\ 				
            DAN(K=2)+MCB  & {61.8 }& { 81.9 }& { 36.0 }& { 50.3 }\\ 				
            DAN(K=3)+MCB  & {62.1 }& { 81.6 }& { 36.6 }& { 50.6 }\\ 
            DAN(K=4)+MCB  & \textbf{62.7}& { 81.9 }& \textbf{ 37.2 }& \textbf{ 50.9}\\ \hline 
    		DAN(K=5)+LQI & {56.1 }& { 78.1 }& { 35.0 }& { 35.2 } \\ 
    		DAN(K=1,Random)+LQI & { 54.3 }& { 77.5 }& { 33.3}& {34.5} \\ \hline
    \end{tabular}
	\vspace{-0.3cm}
\end{table}

\begin{table*}[htb]
	\caption{Analysis network parameter for DAN }
	\vspace{-1.4em}
	\label{tab:dan_param}
    \centering
    \begin{tabular}{l|cccr | c } \hline
    	\textbf{Models}&  \multicolumn{4}{c|}{VQA1.0 Open-Ended (test-dev)}&  HAT val dataset \\ 
    	 \cline{2-6}
    	 & \textbf{All} & \textbf{Yes/No} & \textbf{Number} & \textbf{others} & \textbf{Rank-correlation} \\ \hline 	
    	 	LSTM Q+I+ Attention(LQIA) & { 56.1 }& { 80.3 }& { 37.4 }& {  40.46}& { 0.2142} \\ \hline
    		DAN(K=1)+LQIA  & {59.2 }& { 80.1 }& { 36.1 }& { 46.6}& { 0.2959} \\ 				
    		DAN(K=2)+LQIA  & {59.5 }& { 80.9 }& { 36.6 }& { 47.1 }& { 0.3090} \\ 				
    		DAN(K=3)+LQIA  & {59.9 }& { 80.6 }& { 37.2 }& \textbf{ 47.5 }& { 0.3100}\\ 
    		DAN(K=4)+LQIA  & \textbf{60.2 }& \textbf{ 80.9 }& \textbf{ 37.4 }& { 47.2 } & \textbf{ 0.3206 }\\ \hline 
    		
            DAN(K=1)+MCB  & {64.8 }& { 82.4 }& { 38.1 }& { 54.2}& { 0.3284} \\ 				
            DAN(K=2)+MCB  & {64.8 }& { 82.9 }& { 38.0 }& { 54.3 }& { 0.3298} \\ 				
            DAN(K=3)+MCB  & {64.9 }& { 82.6 }& { 38.2 }& { 54.6 }& { 0.3316}\\ 
            DAN(K=4)+MCB  & \textbf{65.0}& { 83.1 }& \textbf{ 38.4 }& \textbf{ 54.9}& \textbf{ 0.3326} \\ \hline 
    
    		DAN(K=5)+LQIA & {58.1 }& { 79.4 }& { 36.9 }& { 45.7 }& { 0.2157} \\ 
    		DAN(K=1,Random)+LQIA & { 56.4 }& { 79.3 }& { 37.1 }& {  44.6} & {0.2545}\\ \hline
    \end{tabular}
	\vspace{-0.3cm}
\end{table*}

\begin{table*}[htb]
	 	\caption{Analysis network parameter for DCN}
	 	\vspace{-1.4em}
	 	\label{tab:dcn_param}
	 	\begin{center}
		
		\begin{tabular}{l|cccr | c }  \hline 
		 \textbf{Models}&  \multicolumn{4}{c|}{VQA1.0 Open-Ended (test-dev)}&  HAT val dataset \\ 
		 \cline{2-6}
			& \textbf{All} & \textbf{Yes/No} & \textbf{Number} & \textbf{others} & \textbf{Rank-correlation} \\ \hline 
			LSTM Q+I+ Attention(LQIA) & { 56.1 }& { 80.3 }& { 37.4 }& {  40.46} & {  0.2142}\\ \hline
			DCN Add\_v1(K=4)+(LQIA)      & {60.4 }& {81.0 }& { 37.5 }& { 47.1}& {0.3202} \\ 		
			DCN Add\_v2(K=4)+(LQIA) & {60.4 }& {81.2 }& { 37.2 }& { 47.3 } & {0.3215}\\ 
			DCN Mul\_v1(K=4)+(LQIA)      & {60.6 }& {80.9 }& \textbf{ 37.8 }& { 47.9 }& {0.3229} \\ 
			DCN Mul\_v2(K=4)+(LQIA)  & \textbf{60.9 }& \textbf{81.3 }& { 37.5 }& \textbf{ 48.2 }& \textbf{0.3242} \\ \hline
			
            DCN Add\_v1(K=4)+MCB  & {65.1}& { 83.1 }& { 38.5 }& { 54.5 }& {0.3359}\\
            DCN Add\_v2(K=4)+MCB  & {65.2}& { 83.4 }& { 39.0 }& { 54.6 }& {0.3376}\\
            DCN Mul\_v1(K=4)+MCB  & {65.2}& \textbf{ 83.9 }& { 38.7 }& { 54.9 }& {0.3365}\\
            DCN Mul\_v2(K=4)+MCB  & \textbf{65.4}& { 83.8 }& \textbf{ 39.1 }& \textbf{ 55.2 }& \textbf{0.3389}\\ \hline
		\end{tabular}
	\end{center}
	\vspace{-0.7em}
\end{table*}

\subsection{Fusion based Exemplar Approach}\label{subsec:methodvqg}
In this section, the image embedding is concatenated with the question embedding to solve for the answer using a softmax function in a multiple-choice setting instead of attention module for the VQA classification task. We name this model as a Differential Joint Network (DJN). We extend the DJN method for the generative framework (VQG). We name that MDN-Joint method. 
Our (MDN-Joint) method is similar to MDN-attention. The sequence to sequence network has a text sequence as input and output. 

\subsubsection{Fusion Exemplar for VQA: Differential Joint Network (DJN)}
Instead of attention network,  we use a fusion network which concatenates image feature with question feature to predict the final answer as shown in the figure-\ref{fig:MDN} (a). More formally, given an image $x_i$, we obtain an embedding $g_i$ using a CNN that we parameterized through a function $G(x_i, W_c)$ where $W_c$ are the weights of the CNN. Similarly, the question $q_i$ results in a question embedding $f_i$ after passing through an LSTM parameterised using the function $F(q_i, W_l)$ where $W_l$ are the weights of the LSTM. This is similar to part 1 of figure-~\ref{fig:DAN}(a). The output image embedding $g_i$ and question embedding $f_i$ are used in a concatenated network that combines the image and question embeddings and produces an output feature vector $s_i$. The joint mechanism is illustrated in figure-~\ref{fig:MDN}(a). The weights of this network are learned end-to-end using the two losses, a triplet loss and a soft-max classification loss for the answer (shown in part 3 of figure-~\ref{fig:MDN}(a)). The aim is similar to that used in the DAN by using exemplars. The joint loss function used for training is similar to equation-\ref{eq1}.
%
\subsubsection{Fusion Exemplar for VQG: Multimodal Differential Network}\label{subsec:mdn_fuse}
This method is similar to the method discussed in section-\ref{subsec:mdn_attention}.  The main difference is in the mixture module. Section-\ref{subsec:mdn_attention} is attention based mixture module, while in this section-\ref{subsec:mdn_fuse}, we have evaluated  three different approaches for fusion viz., joint, element-wise addition, and Hadamard method. Each of these variants receives image features $g_i$ \& the caption embedding $f_i$, and outputs a fixed dimensional feature vector $s_i$.  
The joint method concatenates $g_{i}$  \& $f_{i}$ and maps them to a fixed length feature vector $s_{i}$ as follows:
\begin{equation}
     s_{i} =  W^{T}_{j} *  \tanh(  W_{ij}  g_{i} \ ^\frown \ (W_{cj}  f_{i} + b_j))
\end{equation}
where $g_{i}$ is  the 4096-dimensional  convolutional feature from the  FC7 layer of pretrained VGG-19 Net~\cite{simonyan_arXiv2014}. $W_{ij}, W_{cj},W_{j}$ are the weights and $b_j$ is the bias for different layers. $^\frown$ is a operator which can be used for concatenation, addition or element wise multiplication operator.  Using this joint operator case we achieve the best score in VQG.

\subsection{Cost function}
\noindent Our objective is to minimize the total loss, which is the sum of cross-entropy loss and triplet loss over all training examples.
  The total loss is: 
  \begin{equation}
    L= \frac{1}{M} \sum^{M}_{i=1} (L_{cross} + \gamma L_{triplet})
\end{equation}
where  $M$ is the total number of samples,$\gamma$ is a constant, which controls both the loss. $L_{triplet}$ is the triplet loss function~\ref{triplet_loss}. $L_{cross}$ is the cross entropy loss between the predicted and ground truth questions and is given by: 
 \begin{equation*}
L_{cross}=\frac{-1}{N}\sum_{t=1}^{N} {y_{t} \texttt{log} P(\hat{q_{t}}|I_i,C_i,{\hat{q_0},..\hat{q_{t-1}}})}
\end{equation*}
where, $N$ is the total number of question tokens, $y_t$ is the ground truth label. The code for  MDN-VQG model is provided \footnote{The project page for MDN-VQG Model is \url{https://badripatro.github.io/MDN-VQG/}}. 

We further extend the model to a differential attention setting where we bring two supporting exemplar embeddings closer and push two contrasting exemplar embeddings far away from the target embedding in attention space. We provide a detailed analysis of this network in the results section in the supplementary material. 

We experimented with ITML based metric learning~\cite{davis_ACM2007} for obtaining exemplars using image features. Surprisingly, the Euclidean KNN-based approach outperforms the ITML one. We also tried random exemplars and a different number of exemplars and found that $k=5$ works best. We provide these results in the supplementary material.

\begin{table}[ht]
	\caption{Open-Ended VQA1.0 accuracy on test-dev}
	\vspace{-1.4em}
	\label{VQA1_accuracy}
	\begin{center}			
		\begin{tabular}{lcccr  } \hline
			
			\textbf{Models} & \textbf{All} & \textbf{Yes/No} & \textbf{Number} & \textbf{others}  \\ \hline 
			LSTM Q+I \cite{VQA}& { 53.7 }& {78.9 }& { 35.2}& { 36.4} \\
		    LSTM Q+I+ Attention(LQIA) & { 56.1 }& { 80.3 }& { 37.4 }& {  40.4} \\
			DPPnet \cite{Noh_CVPR2016}    & { 57.2 }& {80.7 }& { 37.2 }& { 41.7} \\
		 	SMem \cite{Xu_ECCV2016}       & { 58.0 }& {80.9 }& { 37.3 }& { 43.1} \\ 
			SAN \cite{Yang_CVPR2016}      & { 58.7 }& {79.3 }& { 36.6 }& { 46.1 }\\
			QRU(1)\cite{Li_NIPS2016}  & { 59.3 }& {81.0 }& { 35.9 }& { 46.0 }\\\hline
		
			DAN(K=4)+ LQIA & \textbf{60.2 }& { 80.9 }& \textbf{ 37.4 }& \textbf{ 47.2 } \\ \hline
			DMN+\cite{Xiong_arXiv2016} & {60.3 }& { 80.5 }& { 36.8 }& { 48.3 } \\ 
			QRU(2)\cite{Li_NIPS2016}  & { 60.7 }& {82.3 }& { 37.0 }& { 47.7 }\\\hline				
			DCN Mul\_v2(K=4)+LQIA  & \textbf{60.9 }& {81.3 }& \textbf{ 37.5 }& \textbf{ 48.2 } \\ \hline

			HieCoAtt \cite{Lu_NIPS2016} & {61.8}& { 79.7 }& { 38.9 }& { 51.7 } \\
			
			MCB + att \cite{Fukui_arXiv2016} & {64.2}& { 82.2 }& { 37.7 }& { 54.8 } \\
				
			MLB \cite{Kim_ICLR2017} & {65.0}& \textbf{ 84.0 }& { 37.9 }& { 54.7 } \\ \hline
            DAN(K=4)+ MCB  & \textbf{65.0}& { 83.1 }& \textbf{ 38.4 }& \textbf{ 54.9} \\ 
            DCN Mul\_v2(K=4)+MCB  & \textbf{65.4}& { 83.8 }& \textbf{ 39.1 }& \textbf{ 55.2 } \\ \hline
		\end{tabular}
	\end{center}
	\vspace{-1.2em}
\end{table}
\section{Experiments for VQA task}
\label{sec:results}
The experiments have been conducted using the two variants of differential attention that are proposed and compared against baselines on standard datasets. We first analyze the different parameters for the two variants DAN and DCN, that are proposed. We further evaluate the two networks by comparing the networks with comparable baselines and evaluate the performance against the state of the art methods. The main evaluation is conducted to evaluate the performance in terms of correlation of attention with human correlation where we obtain state-of-the-art in terms of correlation with human attention. Further, we observe that its performance in terms of accuracy for solving the VQA task is substantially improved and is competitive with the current state of the art results on standard benchmark datasets. We also analyse the performance of the network on the recently proposed VQA2 dataset. 
\begin{table}[ht]
			\caption{VQA2.0 accuracy on Validation set for DCN and DAN }
			\vspace{-1.4em}
			\label{VQA2_accuracy}
			\begin{center}
				
				\begin{tabular}{lcccr  } \hline
					
					\textbf{Models} & \textbf{All} & \textbf{Yes/No} & \textbf{Number} & \textbf{others}  \\ \hline 
					SAN-2           & {52.82 }& {- }& { - }& { - }\\  \hline 
					DAN(K=1) +LQIA        &\textbf {52.96}& \textbf{70.08 }& \textbf{34.06 }& \textbf{ 44.20} \\  \hline
					
					DCN Add\_v1(K=1)+LQIA     & {53.01 }& { 70.13 }& { 33.98 }& { 44.27} \\ 
					DCN Add\_v2(K=1) +LQIA& {53.07 }& { 70.46 }& { 34.30 }& { 44.10 } \\ 					
					DCN Mul\_v1(K=1) +LQIA    & {53.18 }& { 70.24 }& { 34.53 }& { 44.24 } \\ 					
					DCN Mul\_v2(K=1)+LQIA & \textbf{53.26}& \textbf{ 70.57 }& \textbf{ 34.61}& \textbf{ 44.39 } \\ \hline
					
					DCN Add\_v1(K=4)+MCB & {65.30 }& { 81.89 }& { 42.93 }& { 55.56 } \\ 
					DCN Add\_v2(K=4)+MCB & {65.41 }& { 81.90 }& { 42.88 }& { 55.99 } \\ 					
					DCN Mul\_v1(K=4)+MCB & {65.52 }& { 82.07 }& { 42.91 }& { 55.97 } \\ 					
					DCN Mul\_v2(K=4)+MCB & \textbf{65.90}& \textbf{ 82.40 }& \textbf{ 43.18}& \textbf{ 56.81 } \\ \hline

				\end{tabular}
				
			\end{center}
			
\end{table}

\subsection{Analysis of Network Parameters for VQA}
\label{sec:param_analysis}
In the proposed DAN network, we have a dependency on the number of k-nearest neighbors that should be considered. We observe in table-~\ref{tab:dan_param}, that is using 4 nearest neighbors in the triplet network, we obtain the highest correlation with human attention as well as accuracy using VQA-1 dataset. We, therefore, use 4 nearest neighbors in our experiments. We observe that increasing nearest neighbors beyond 4 nearest neighbors results in a reduction in accuracy. Further, even using a single nearest neighbor results in substantial improvement that is marginally improved as we move to 4 nearest neighbors. 

We also evaluate the effect of using the nearest neighbors as obtained through a baseline model~\cite{{VQA}} versus using a random assignment of supporting and opposing exemplar. We observe that using DAN with a random set of nearest neighbors decreases the performance of the network. While comparing the network parameters, the comparable baseline we use is the basic model for VQA using LSTM and CNN \cite{{VQA}}. This, however, does not use attention, and we evaluate this method with attention. With the best set of parameters, the performance improves the correlation with human attention by 10.64\%. We also observe that correspondingly the VQA performance improves by 4.1\% over the comparable baseline. We further then incorporate this model with the model from MCB \cite{Fukui_arXiv2016}, which is a state of the art VQA model. This further improves the result by 4.8\% more on VQA and a further increase in correlation with human attention by 1.2\%. 
     
     
 In the proposed DCN network, we have two different configurations, one where we use the add module (DCN-add) for adding the differential context feature and one where we use the (DCN-mul) multiplication module for adding the differential context feature. We further have a dependency on the number of k-nearest neighbors for the DCN network as well. This is also considered. We next evaluate the effect of using a fixed scaling weight (DCN\_v1) for adding the differential context feature against learning a linear scaling weight (DCN\_v2) for adding the differential context feature. All these parameter results are compared in table-~\ref{tab:dcn_param}.

 As can be observed from table-~\ref{tab:dcn_param}, the configuration that obtains maximum accuracy on VQA dataset \cite{VQA} and in correlation with human attention is the version that uses multiplication with learned weights and with 4 nearest neighbors being considered. This results in an improvement of 11\% in terms of correlation with human attention and 4.8\% improvement in accuracy on the VQA-1 dataset \cite{VQA}. We also observe that incorporating DCN with MCB \cite{Fukui_arXiv2016} further improves the results by 4.5\% further on the VQA dataset and results in an improvement of 1.47\% improvement in correlation with attention. These configurations are used in comparison with the baselines. 
 \begin{table}[ht]
	\caption{Rank Correlation on HAT Validation Dataset for DAN and DCN}
	\vspace{-1.4em}
	\label{HAT_rank_correlation}
	\begin{center}		
		\begin{tabular}{ lc } \hline
			
			\textbf{Models} & \textbf{Rank-correlation} \\ \hline 
			LSTM Q+I+ Attention(LQIA)	 			& {0.214 $\pm$ 0.001} \\
			SAN\cite{Das_EMNLP2016} 		& {0.249 $\pm$  0.004} \\			
			HieCoAtt-W\cite{Lu_NIPS2016}	& {0.246 $\pm$ 0.004} \\		
			HieCoAtt-P \cite{Lu_NIPS2016}	& {0.256 $\pm$ 0.004}\\
			HieCoAtt-Q\cite {Lu_NIPS2016}	& {0.264 $\pm$ 0.004}\\
			MCB + Att.	& {0.279 $\pm$ 0.004} \\\hline
			DAN (K=4) +LQIA	 					& { 0.321$\pm$ 0.001} \\
			DCN Mul\_v2(K=4) +LQIA				& \textbf{ 0.324$\pm$ 0.001}  \\ \hline
            DAN (K=4) +MCB & { 0.332$\pm$ 0.001} \\
            DCN Mul\_v2(K=4) +MCB & \textbf{ 0.338$\pm$ 0.001}  \\ \hline
			
			Human \cite{Das_EMNLP2016}   	& { 0.623 $\pm$ 0.003 } \\ \hline
		\end{tabular}	
	\end{center}
\end{table}

\subsection{Comparison with baseline and state of the art for VQA}
We obtain the initial comparison with the baselines on the rank correlation on human attention (HAT) dataset \cite{Das_EMNLP2016} that provides human attention by using a region deblurring task while solving for VQA. Between humans, the rank correlation is 62.3\%. The comparison of various state-of-the-art methods and baselines are provided in table-~\ref{HAT_rank_correlation}. The baseline we use \cite{{VQA}} is the method used by us for obtaining exemplars. This uses a question embedding using an LSTM and an image embedding using a CNN. We additionally consider a variant of the same that uses attention. We have also obtained results for the stacked attention network~\cite{Yang_CVPR2016}. The results for the Hierarchical Co-Attention work~\cite{Lu_NIPS2016} are obtained from the results reported in Das {\it et al.}~\cite{Das_EMNLP2016}. We observe that in terms of rank correlation with human attention we obtain an improvement of around 10.7\% using DAN network (with 4 nearest neighbors) and using DCN network (4 neighbors with multiplication module and learned to scale weights) we obtain an improvement of around 11\% over the comparable baseline. We also obtain an improvement of around 6\% over the Hierarchical Co-Attention work \cite{Lu_NIPS2016} that uses co-attention on both images and questions. Further incorporating MCB improves the results for both DAN and DCN resulting in an improvement of 7.4\% over Hierarchical co-attention work and 5.9\% improvement over MCB method. However, as noted by \cite{Das_EMNLP2016}, using a saliency based method  \cite{Judd_ICCV2009} that is trained on eye tracking data to obtain a measure of where people look in a task independent manner results in more correlation with human attention (0.49). However, this is explicitly trained using human attention and is not task dependent. In our approach, we aim to obtain a method that can simulate human cognitive abilities for solving tasks.
 
 	\begin{figure}[ht]
	\centering
	\includegraphics[width=0.4\textwidth]{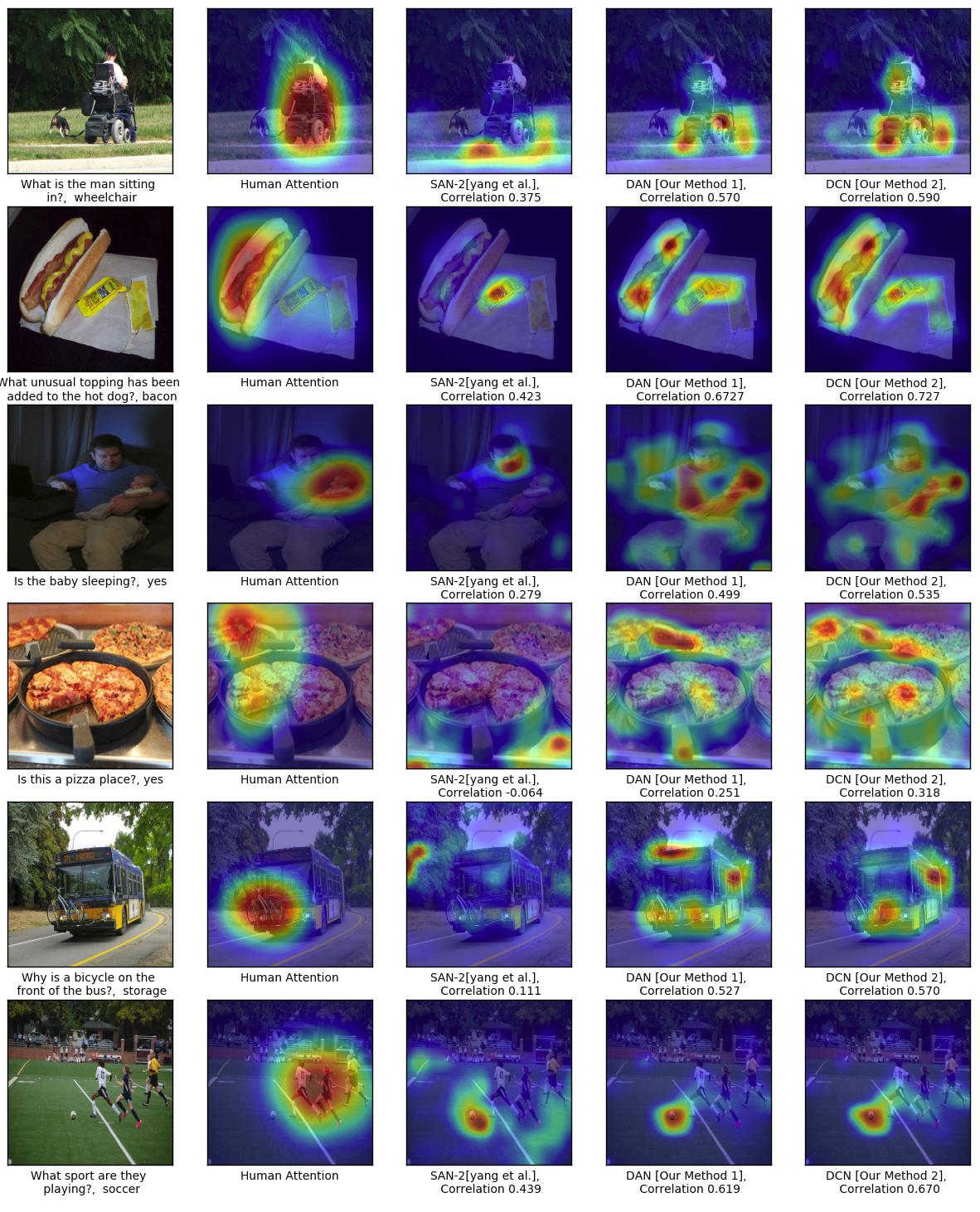}
	\vspace{-0.8em}
	\caption{ In this figure, the first column indicates target question and corresponding image, second column indicates reference human attention map in HAT dataset, third column refer to generated attention map for SAN, fourth column refers to rank correlation of our DAN model and final column refers to rank correlation for our DCN model.}
	\label{fig:att}
\end{figure}

We next evaluate the different baseline and state of the art methods on the VQA dataset \cite{VQA} in table-~\ref{VQA1_accuracy}. There have been a number of methods proposed for this benchmark dataset for evaluating the VQA task. Among the notable different methods, the Hierarchical Co-Attention work \cite{Lu_NIPS2016} obtains 61.8\% accuracy on the VQA task, the dynamic parameter prediction \cite{Noh_CVPR2016} method obtains 57.2\%, and the stacked attention network \cite{Yang_CVPR2016} obtains 58.7\% accuracy. We observe that the differential context network performs well outperforming all the image based attention methods and results in an accuracy of 60.9\%. This is a strong result, and we observe that the performance improves across different kinds of questions. Further, on combining the method with MCB, we obtain improved results of 65\% and 65.4\% using  DAN and DCN, respectively improving over the results of MCB  by 1.2\%. This is consistent with the improved correlation with human attention that we observe in table-~\ref{HAT_rank_correlation}.
%
%
%

We next evaluate the proposed method on a recently proposed VQA-2 dataset \cite{Goyal_CVPR2017}. The aim of this new dataset is to remove the bias in different questions. It is a more challenging dataset as compared to the previous VQA-1 dataset \cite{VQA}. We provide a comparison of the proposed DAN and DCN methods against the stacked attention network (SAN) \cite{Yang_CVPR2016} method. As can be observed in table-~\ref{VQA2_accuracy}, the proposed methods obtain improved performance over a strong stacked attention baseline. We observe that our proposed methods are also able to improve the result over the SAN method. DCN with 4 nearest neighbors, when combined with MCB, obtains an accuracy of 65.90\%.
		

\begin{figure}[ht]
\vspace{-0.3cm}
	\centering
	\includegraphics[width=0.485\textwidth]{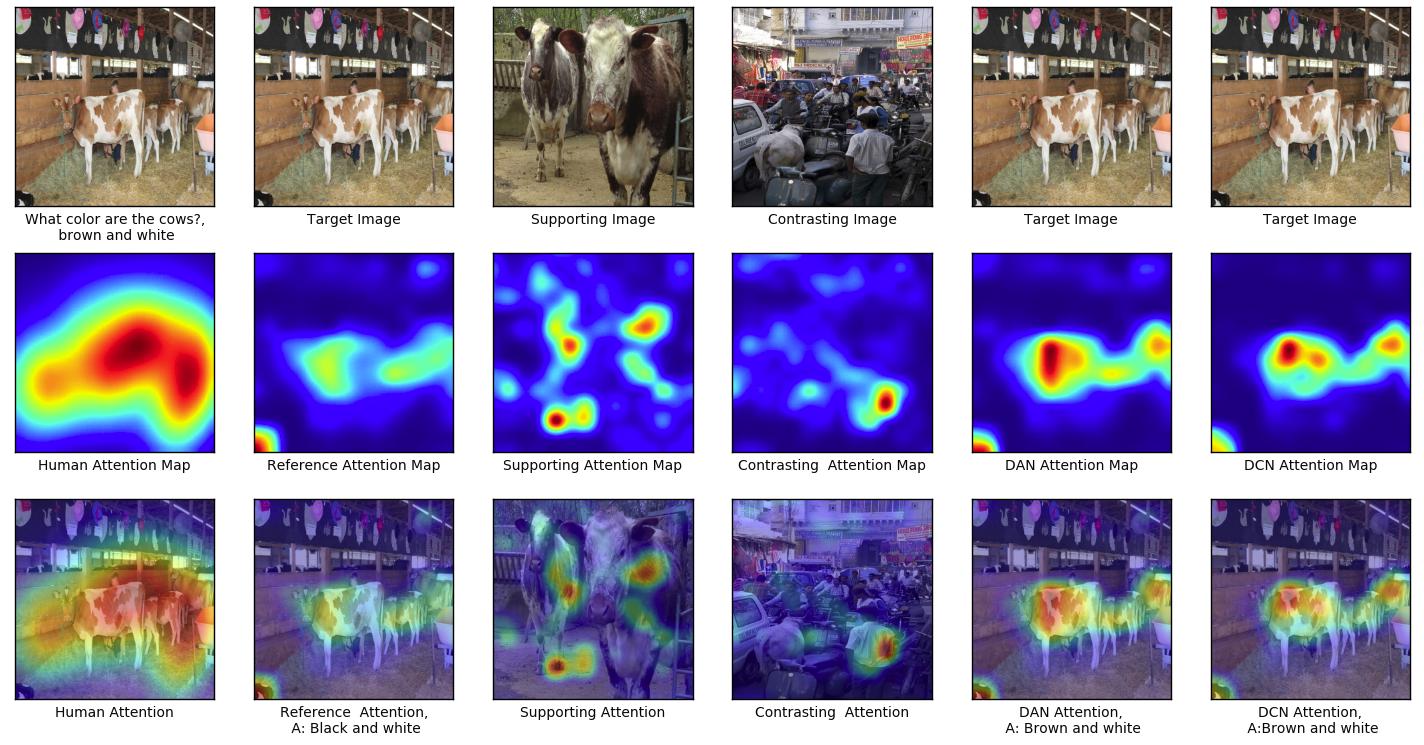}
	\vspace{-1.5em}
	\caption{ In this figure, the first	row indicates the given target image, supporting image and opposing image. second row indicates the attention map for human\cite{Das_EMNLP2016}, reference attention map, supporting attention map , opposing attention map, DAN and DCN attention map respectively. Third row generates result by applying attention map on corresponding images.}
	\label{fig:attention}
\end{figure}		
\subsection{Attention Visualization}
The main aim of the proposed method is to obtain improved attention that correlates better with human attention. Hence we visualize the attention regions and compare them. In attention visualization, we overlay the attention probability distribution matrix, which is the most prominent part of a given image based on the query question. The procedure followed is the same as that followed by Das {\it et al.} \cite{Das_EMNLP2016}. We provide the results of the attention visualization in figure~\ref{fig:att}. We obtain significant improvement in attention by using DCN as compared to the SAN method~\cite{Yang_CVPR2016}. Figure~\ref{fig:attention} provides how the supporting and opposing attention map helps to improve the reference attention using DAN and DCN.  We have provided more results for attention map visualization on the project website \footnote{project website:\url{https://badripatro.github.io/DVQA/} }.
	

	

\subsection{How important are the supporting and contrasting exemplar?}
\begin{figure}[ht]
	\includegraphics[width=0.50\textwidth]{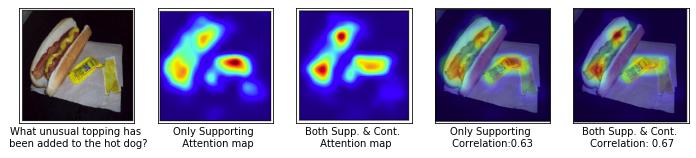}
	\vspace{-1.5em}
	\caption{Importance of Supporting exemplar vs both. the first column in the figure indicates about image and corresponding question, the second  and third term indicates attention map for supporting exemplar and both supporting and opposing exemplar. The fourth and fifth column gives the value of rank correlation for supporting and both.}
	\label{fig:a1}
\end{figure}

 We carried out an experiment by considering only the supportive exemplar in triplet loss mentioned in equation-~\ref{eq2} and obtained consistent results, as shown in figure-~\ref{fig:a1}. From the rank correlation result, we can conclude that, if we use only the supportive exemplar, we can obtain significant gain. However, this is improved by considering both supporting and opposing exemplars. The quantitative results for this ablation analysis are shown in table-~\ref{DAN_rank_correlation}, which provides the rank correlation on HAT Validation Dataset.

  \begin{table}[ht]
	\caption{Rank Correlation for only Supporting Exemplar}
	\vspace{-1.4em}
	\label{DAN_rank_correlation}
	\begin{center}		
		\begin{tabular}{ lc } \hline
			\textbf{Models} & \textbf{Rank-correlation} \\ \hline 
			DAN (K=4) +LQIA	 & { 0.312 $\pm$ 0.001} \\
            DAN (K=4) +MCB & { 0.320 $\pm$ 0.001} \\ \hline 
		\end{tabular}	
	\end{center}
	\vspace{-1.2em}
\end{table}

	\begin{table}[ht]
		\caption{Rank Correlation with Human attention maps on HAT Validation Dataset for DAN and DCN}
		\vspace{-1.4em}
		\label{HAT_rank_correlation1}
		\begin{center}		
			\begin{tabular}{ lcccr } \hline
				
				\textbf{Models}  & \textbf{Map1} & \textbf{ Map2} & \textbf{Map3 }& \textbf{Map} \\ \hline 
				DAN (K=1)  & {0.3147}& { 0.2772}& { 0.2958}& { 0.2959} \\ 
				 
				DAN (K=2)  & {0.3280}& { 0.2933}& { 0.3057}& { 0.3090} \\ 
				DAN (K=3)  & {0.3297}& { 0.2947}& { 0.3058}& { 0.3100} \\ 
				DAN (K=4)  &\textbf {0.3418}&\textbf{ 0.3060}	&\textbf { 0.3133}& \textbf{ 0.3206} \\ \hline
						
				DCN Add\_v1(K=1)      & { 0.3172}& { 0.2783}& { 0.2968}	& { 0.2974} \\ 
				DCN Add\_v2(K=1)  & { 0.3186}& { 0.2812}& { 0.2993}	& { 0.2997} \\ 
				DCN Mul\_v1(K=1)      & {0.3205}&  { 0.2847}& { 0.3023}&  { 0.3025} \\
				DCN Mul\_v2(K=1)  & \textbf{ 0.3227}& \textbf{ 0.2871}& \textbf{ 0.3059}	& \textbf{ 0.3052} \\ \hline

				DCN Add\_v1(K=4)      & {0.3426}& {0.3058}&{ 0.3123}& { 0.3202} \\ 			 	
				DCN Add\_v2(K=4)  & {0.3459}&{ 0.3047}&{ 0.3140}& { 0.3215} \\ 
				DCN Mul\_v1(K=4)      & {0.3466}& {0.3059}&{ 0.3163}& { 0.3229} \\
				DCN Mul\_v2(K=4)  & \textbf{0.3472}&\textbf{ 0.3068}	& \textbf{ 0.3187}& \textbf{ 0.3242}  \\ \hline
				DAN (K=1,Random )  & {0.1238}&{ 0.1070}	& { 0.1163}& { 0.1157} \\ 
				DAN (K=5)  & {0.2634}&{ 0.2412}	& { 0.2589}& { 0.2545} \\
			\hline
					
			\end{tabular}	
		\end{center}
		\vspace{-1.4em}
	\end{table}

\subsection{Comparison of DAN and DCN with Human attentions for VQA}
    We compare the attention probability of our models (DAN or DCN) with Human attention ~\cite{Das_EMNLP2016} provided by HAT for the validation set using rank correlation, as shown in table-~\ref{HAT_rank_correlation1}. The HAT validation dataset contains three human-annotated attention maps per question. Table-~\ref{HAT_rank_correlation1} contains a comparison of rank correlation with respect to all three human attention maps. The first attention map gives better accuracy than the other two. Finally, we take an average of three as a rank correlation of a particular model. We can observe that all of our model attention maps have a positive rank correlation with human attention.
    
    The first block of the table-~\ref{HAT_rank_correlation1} shows rank correlation results for different variants of DAN for k=2 for two supporting and opposing pairs and similarly for k=3,4 for three and four supporting and opposing pair. We observe that the rank correlation result increases by increasing the number of supporting and opposing pairs. But when K=5 and more, the correlation decreases. Also, for random selection of supporting and opposing pairs, the value is decreased, as mentioned in the last block of the table-~\ref{HAT_rank_correlation1}. The second and third block of the table shows the rank correlation for two types of DCN, i.e., DCN Add and DCN Mul. Each type of network has two different methods for training; one is fixed scaling weights, i.e., DCN Mul, and the second one is learn-able scaling weights, i.e., DCN Mul\_v1. The statistics of rank correlation in this table indicate that learnable scaling weights perform better than fixed weights. Also, we observed that the multiplication network performs better than an addition network in case of a differential context. We did experiments for K=1,2,3,4, but this table only shows the results of K=1 and K=4 for the number of nearest and farthest neighbors selections.

 \begin{table}[ht]
\scriptsize
\centering
\caption{Analysis of variants of our proposed method on VQG-COCO dataset as mentioned in supplementary section and different ways of getting a joint embedding (Attention (AtM), Hadamard (HM), Addition (AM) and Joint (JM) method as given in section-~\ref{mixture_model}) for each method. Refer section~\ref{ablation_analysis} for more details. B1 is BLEU1.
}
\begin{tabular}{|l|l|cccc|}
\hline \bf Emb. & \bf Method & \bf B1 & \bf METEOR & \bf ROUGE & \bf CIDEr \\ \hline 

Tag & AtM & 22.4 &8.6 & 22.5 & 20.8\\ 
Tag & HM & \textbf{24.4} &\textbf{10.8}& \textbf{24.3} & \textbf{55.0}\\
Tag & AM & 24.4 &10.6& 23.9 & 49.4\\
Tag& JM &22.2 &10.5 & 22.8 &50.1 \\ \hline

PlaceCNN & AtM &24.4  &10.3 &24.0 &51.8 \\
PlaceCNN & HM &24.0  &10.4 &24.3 &49.8 \\
PlaceCNN & AM & 24.1 & 10.6&24.3 &51.5 \\
PlaceCNN & JM &\textbf{25.7}  &\textbf{10.8 }&\textbf{24.5} &\textbf{56.1}  \\ \hline

Diff. Img & AtM & 20.5 &8.5 & \textbf{24.4} & 19.2\\ 
Diff. Img & HM& 23.6 &8.6 & 22.3 & 22.0\\
Diff. Img & AM & 20.6 &8.5 & 24.4 & 19.2\\ 
Diff. Img & JM & \textbf{30.4} & \textbf{11.7} & 22.3 & \textbf{22.8}\\ \hline

MDN & AtM & 22.4 &8.8 & 24.6 & 22.4\\ 
MDN & HM & 26.6 &12.8 & 30.1 & 31.4\\
MDN & AM & 29.6 &15.4 & 32.8 & 41.6\\ 
MDN\textbf{(Ours)} & JM & \textbf{36.0}&\textbf{23.4}&\textbf{41.8}& \textbf{50.7}\\\hline

\end{tabular}
\label{score_tab_1_vqg}
\vspace{-0.5cm}
\end{table}

\section{Experiment for VQG task}
We evaluate our proposed MDN method in the following ways: First, we evaluate it against other variants (Image exemplar method, Addition model, multiplication model) described in table-\ref{score_tab_1_vqg}. Second, we further compare our network with state-of-the-art methods for VQA 1.0 and VQG-COCO dataset as mentioned in table ~\ref{score_tab_2_vqg} and ~\ref{score_tab_3_vqg} respectable.
 We further consider the statistical significance for the various ablations as well as the state-of-the-art models,user study to gauge human opinion on naturalness of the generated question, analyze the word statistics, dataset and evaluation methods for VQG task are present int he supplementary. 

\begin{table}[ht]
\scriptsize
\caption{\label{score_tab_2_vqg}State-of-the-Art comparison on VQA-1.0 Dataset. The first block consists of the state-of-the-art results, second block refers to the baselines mentioned in section~\ref{baseline_sota}, third block provides the results for the variants of mixture module present in section~\ref{mixture_model}.}
\centering
\begin{tabular}{|l|cccc|}
\hline \bf Methods & \bf BLEU1  & \bf METEOR & \bf ROUGE &  \bf CIDEr \\ \hline
Sample(\cite{Yang_arXiv2015}) & 38.8  & 12.7 & 34.2 & 13.3 \\
Max(\cite{Yang_arXiv2015}) &  59.4  & 17.8 & 49.3 & 33.1\\ \hline
Image Only & 56.6  & 15.1  & 40.0 & 31.0\\
Caption Only & 57.1  & 15.5  & 36.6 & 30.5\\ \hline
MDN-Attention   & 60.7  & 16.7  & 49.8 & 33.6\\
MDN-Hadamard  & 61.7 & 16.7  & 50.1 & 29.3 \\
MDN-Addition  & 61.7  & 18.3  & 50.4 & \textbf{42.6} \\
MDN-Joint (\bf Ours)& \textbf{65.1}  & \textbf{22.7}  & \textbf{52.0} & 33.1\\
\hline
\end{tabular}
\end{table}
\subsection{Ablation Analysis}\label{ablation_analysis}
The results for the VQG-COCO test set are given in table~\ref{score_tab_1_vqg}. In this table, every block provides the results for one of the variations of obtaining the embeddings and different ways of combining them. We observe that the Joint Method (JM) of combining the embeddings works the best in all cases except the Tag Embeddings. Among the ablations, the proposed MDN method works way better than the other variants in terms of BLEU, METEOR, and ROUGE metrics by achieving an improvement of 6\%, 12\%, and 18\% in the scores respectively over the best another variant. 

\subsection{Baseline and State-of-the-Art}\label{baseline_sota}
The comparison of our method with various baselines and state-of-the-art methods is provided in table~\ref{score_tab_2_vqg} for VQA 1.0 and table~\ref{score_tab_3_vqg} for VQG-COCO dataset. The comparable baselines for our method are the image based and caption based models in which we use either only the image or the caption embedding and generate the question. In both the tables, the first block consists of the current state-of-the-art methods on that dataset, and the second contains the baselines. We observe that for the VQA dataset, we achieve an improvement of 8\% in BLEU and 7\% in METEOR metric scores over the baselines, whereas, for the VQG-COCO dataset, this is 15\% for both the metrics. We improve over the previous state-of-the-art~\cite{Yang_arXiv2015} for the VQA dataset by around 6\% in the BLEU score and 10\% in METEOR score. In the VQG-COCO dataset, we improve over~\cite{mostafazadeh2016generating} by 3.7\% and~\cite{jain2017creativity} by 3.5\% in terms of METEOR scores.

\begin{table}[ht]
\scriptsize
\centering
\begin{tabular}{|l|lccc|}
\hline \bf Context &  \bf BLEU1 & \bf METEOR & \bf ROUGE & \bf CIDEr \\ \hline 

Natural \cite{mostafazadeh2016generating} & 19.2 & 19.7  &- & -  \\
Creative \cite{jain2017creativity} & {35.6} & 19.9 & - & - \\ \hline
Image Only &  20.8  &  8.6  & 22.6 & 18.8\\
Caption Only & 21.1 & 8.5  & 25.9 & 22.3\\\hline
Tag-Hadamard & {24.4} &{10.8}& {24.3} & {55.0}\\ 
Place CNN-Joint & {25.7}  &{10.8}&{24.5} &\textbf{56.1} \\
Diff.Image-Joint& 30.4 & {11.7} & 26.3 & {38.8}\\
MDN-Joint (\bf Ours)& \textbf{36.0}&\textbf{23.4}&\textbf{41.8}& 50.7\\\hline
Humans \cite{mostafazadeh2016generating} & \textbf{86.0}&\textbf{60.8}&\textbf{-}& -\\\hline
\end{tabular}
\caption{\label{score_tab_3_vqg}State-of-the-Art (SOTA) comparison on VQG-COCO Dataset. The first block consists of the SOTA results, second block refers to the baselines mentioned in section~\ref{baseline_sota}, third block shows the results for the best method for different ablations mentioned in table~\ref{score_tab_1_vqg}. }
\vspace{-2em}
\end{table}

\section{Discussion}
In this section, we further discuss different aspects of our method that are useful for understanding the method in more detail.

We first consider how exemplars improve attention. In differential attention networks, we use the exemplars and train them using a triplet network. It is known that using a triplet (\cite{Hoffer_Springer2015} and earlier by \cite{Frome_ICCV2007}), that we can learn a representation that accentuates how the image is closer to the supporting exemplar as against the opposing exemplar. The attention is obtained between the image and language representations. Therefore the improved image representation helps in obtaining an improved attention vector. In DCN, the same approach is used with the change that the differential exemplar feature is also included in the image representation using projections. More analysis in terms of understanding how the methods qualitatively improve attention is included in the project website. 

We next consider whether improved attention implies improved performance. In our empirical analysis, we observed that we obtain improved attention {\it and} improved accuracies in the VQA task. While there could be other ways of improving performance on VQA (as suggested by MCB\cite{Fukui_arXiv2016} ), these can be additionally incorporated with the proposed method, and these do yield improved performance in VQA.

Lastly, we consider whether the image (I) and question embedding (Q) are both relevant. We had considered this issue and had conducted experiments by considering I only, by considering Q only, and by considering nearest neighbor using the semantic feature of both Q \& I. We had observed that the Q \&I embedding from the baseline VQA model performed better than other two. Therefore we believe that both contribute to the embedding. 

\section{Conclusion}
\vspace{-0.3em}
In this paper, we propose a generic exemplar based method and explore various variants of the method for two tasks, VQA and VQG. In all the variants that consider attention or fusion using various settings, we observe that including an exemplar module always consistently results in improved attention and accuracy results for the two tasks. This suggests that an exemplar based approach is a useful technique that can be adopted in various vision and language tasks to aid the models to generalize further. We also observed that instead of only incorporating supporting exemplars, also including opposing exemplars aided us in our various tasks. In the future, we would like to consider various such modules which can be used to improve the performance for challenging vision and language based tasks. 

{\small
\bibliographystyle{IEEEtran}
\bibliography{reference}
}

\appendix
\begin{appendices}
\section*{Supplementary Material}\label{sec:Supplementary}
In this section we provide various  experiment details for Visual Question Answering (VQA) task as described in section-\ref{exp_vqa}and Visual Question Generation(VQG) task in section-\ref{exp_vqg}. We provide more ablation analysis for the VQG task, as mentioned in section-\ref{disc} and training and experimental setup, as described in section-\ref{sec:exp}. variation of our proposed model is described in section-\ref{subsec:variants}.

\section{Experiment on VQA task}\label{exp_vqa}
In this section, we provide following details as pseudo-code for obtaining differential attention map as mentioned in section-\ref{ps}, analysis on the contribution of each term in DCN for VQA as described in section-\ref{dcn_contri}, visualizing attention with Supporting and Opposing  Exemplar as described in section\ref{av_1}, attention visualization with a different set of human attention maps present in the validation set of HAT dataset as described in section-\ref{av_2}, more qualitative results are present in section-\ref{av_3}. In this section, we provide a computation of gradient for triplet loss, as mentioned in section-\ref{gradient_cal}.  Data set details and evaluation methods on various dataset are mentioned on section-\ref{dataset_hat_vqa} and \ref{train_eval}.
\subsection{Pseudo Code for Differential Attention Mechanism}\label{ps}
In this section, we provide the pseudo-code -~\ref{algo_attention}  for differential attention mechanism illustrates the core idea of differential attention based on exemplar theory, also provides the dimensions of each input and outputs.

\begin{algorithm}[ht]
\small
		\caption{Differential Attention Mechanism for (VQA)}\label{algo_attention}
		\begin{algorithmic}[1]
		
		    \Procedure{:Model}{$g_{i}$,$g_{s}$,$g_{c}$,$f_{i}$}
			\BState\emph{ Compute Attention Maps}:
			\State $s_i=ATTENTION\_MAP(g_{i},f_{i}) $, $s_i \in \mathcal{R}^{1 \times 196}$
			\State $s^+_i=ATTENTION\_MAP(g_{s},f_{i})$, $s^+_i \in \mathcal{R}^{1 \times 196}$
			\State $s^-_i=ATTENTION\_MAP(g_{c},f_{i})$, $s^-_i \in \mathcal{R}^{1 \times 196}$
			
			\BState\emph{If DAN Model}:
            \State $Loss\_Triplet= triplet\_loss(s_i,s^+_i,s^-_i)$
		    \State $P_{att}= s_i$, $P_{att} \in \mathcal{R}^{1 \times 196}$

            \BState \emph{If DCN Model}:
			\State Compute Context: $r^+_i,r^-_i$ as in eq-3 \& eq-4 
			\State $d_i=s_i \sym  tanh(W_1 r^+_i- W_2 r^-_i )$,  $d_i \in \mathcal{R}^{1 \times 196}$
			\State $P_{att}= d_i$,  $P_{att} \in \mathcal{R}^{1 \times 196}$
            \State [ If \textbf{DCN-mul}, then $\sym$ = {*}. If \textbf{DCN-add}, then  $\sym$ = {+} ]
            \BState \emph{Compute Img \& Ques Attention  }:
			\State $V_{att}= \sum_{i}{P_{att}(i)}{G_{imgfeat}(i)}$,  $V_{att} \in \mathcal{R}^{1 \times 512}$
			\State $A_{att}= V_{att} + f_{i}$,   $A_{att} \in \mathcal{R}^{1 \times 512}$
			\State $Ans=softmax({W_A} A_{att} + b_A)$, $Ans \in \mathcal{R}^{1 \times 1000}$
			\EndProcedure
			\State {------------------------------------------------------------------------}		
			
			\Procedure{:Attention\_map}{$g_{i}$,$f_{i}$}
			\State $g_{i}$:Image feature, $g_{i} \in \mathcal{R}^{14 \times 14 \times 512}$
			\State $f_{i}$: Question feature, $f_{i} \in \mathcal{R}^{1 \times 512}$
			
			\BState \emph{Match dimension}:
			\State $G_{imgfeat}$:Reshape  $g_{i}$ to $196 \times 512$ :$reshape(g_{i})$
			\State  $F_{quesfeat}$: Replicate $f_i$ to 196 times: $clone(f_{i})$ 
			\BState \emph{Compute Attention Distribution}:
			\State $h_{att}= tanh({W_I}{G_{imgfeat}} \Pexp ({W_Q}{F_{quesfeat}}+{b_q}))$
			\State $P_{vec}= softmax({W_P}{h_{att}}+{b_P})$, $P_{vec} \in \mathcal{R}^{1 \times 196}$
			\State Return $P_{vec}$
			\EndProcedure
		\end{algorithmic}
		\end{algorithm}
\subsection{Contribution of each term in DCN for VQA}\label{dcn_contri}
\begin{figure}[ht]
	\includegraphics[width=0.4698\textwidth]{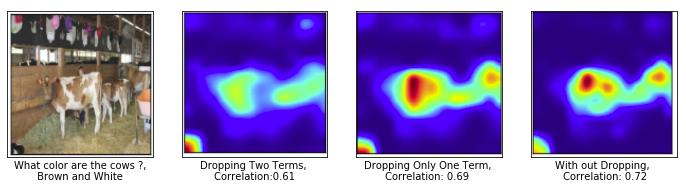}
	\vspace{-1.6em}
	\caption{ Ablation Results for Dropping terms in equations 3 and 4. The first column indicates the target image and its question; The second column provides the attention map \& rank correlation by dropping $2^{nd}$ in equation 3 \& $i^{st}$  term in equation 4. The third column gives the attention map \& rank correlation by dropping only  $i^{st}$  term in equation 4. The final column provides the attention map \& rank correlation by considering everything in both the equation.}
	\label{fig:a2}
	\vspace{-1.6em}
\end{figure}

 We carried out an experiment by dropping the vector projection of $s_i^-$ on $s_i$ term in the supporting context $r_i^+$ as mentioned in equation-3 (in main paper) and the vector rejection of $s_i^+$ on $s_i$ term in opposing context $r_i^-$ as mentioned in equation-4 (in main paper) and obtained consistent result as shown in  figure-~\ref{fig:a2}. The contribution of these terms in the corresponding equations is very small. The quantitative results for this ablation analysis are shown in table-~\ref{DCN_rank_correlation}, which provides the rank correlation on the HAT validation dataset.
\begin{table}[ht]
	\caption{Rank Correlation by Dropping various terms in DCN}
	\vspace{-1.6em}
	\label{DCN_rank_correlation}
	\begin{center}		
		\begin{tabular}{ lc } \hline
			\textbf{Models} & \textbf{Rank-correlation} \\ \hline 
			DCN Mul\_v2(K=4) +LQIA	& \textbf{ 0.319$\pm$ 0.001} \\
            DCN Mul\_v2(K=4) +MCB & \textbf{ 0.3287$\pm$ 0.001}  \\ \hline
		\end{tabular}	
	\end{center}
\end{table}

	\begin{figure*}[ht]
	\centering
	\includegraphics[width=0.95\textwidth]{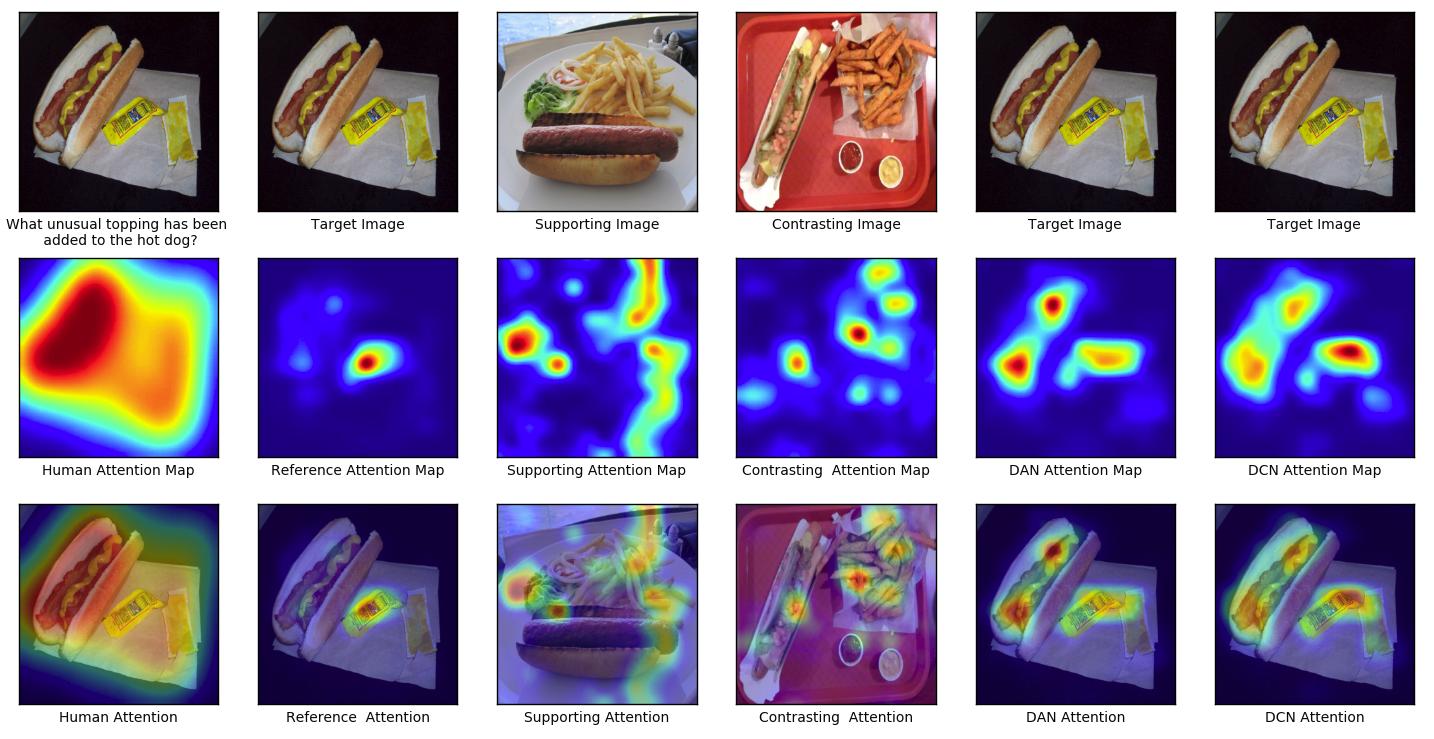}
	\vspace{-1em}
	\caption{In this figure, the first raw indicates the given target image, supporting image, and opposing image. The second raw indicate the attention map for human\cite{Das_EMNLP2016}, reference attention map, supporting attention map, opposing attention map, DAN, and DCN attention map, respectively. Third raw generate results by applying the attention map on corresponding images.}
	\label{fig:res6}
\end{figure*}

\begin{figure*}[ht]
	\centering
	\includegraphics[width=0.95\textwidth]{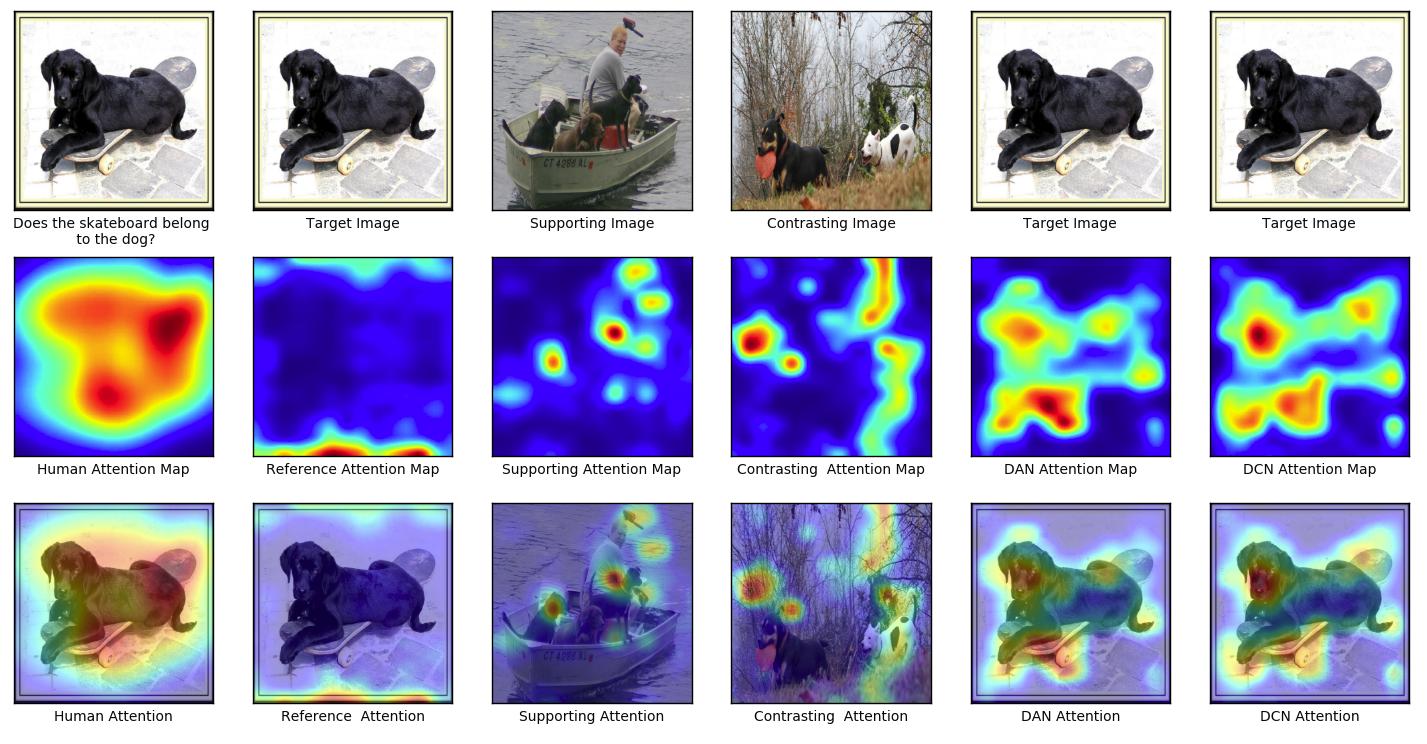}
	\vspace{-1.4em}
	\caption{In this figure, the first raw indicates the given target image, supporting image, and opposing image. The second raw indicate the attention map for human\cite{Das_EMNLP2016}, reference attention map, supporting attention map, opposing attention map, DAN, and DCN attention map, respectively. Third raw generate results by applying the attention map on corresponding images.}
	\label{fig:res7}
\end{figure*}



	\subsection{Attention Visualization for VQA}
    The main aim of the proposed method is to obtain improved attention that correlates better with human attention. Hence we visualize the attention regions and compare them. In attention visualization we overlay the attention probability distribution matrix, which is the most prominent part of a given image based on the query question. The procedure followed is the same as that followed by Das {\it et al.} \cite{Das_EMNLP2016}. 
	

	
	
\subsubsection{ Attention Visualization DAN and DCN with Supporting and Opposing  Exemplar for VQA } \label{av_1}
The first row of figure-~\ref{fig:res6} indicates the target image along with a supporting and opposing image.  The second row provides human attention map, reference, supporting, opposing, DAN, and DCN attention map, respectively. The third row gives corresponding attention visualization for all the images. We can observe that from the given the target image and question: "what unusual topping has been added to the hot dog" , the reference model provides attention map($3^{rd}$ row, $2^{nd}$ column of figure-~\ref{fig:res6}) somewhere in the yellow part which is different from the ground truth human attention map ($3^{rd}$ row, $1^{st}$ column of figure-~\ref{fig:res6}). With the help of supporting and contrasting exemplar attention map($3^{rd}$ row, $3^{rd}$ \& $4^{th}$ column of figure-~\ref{fig:res6}), the reference model attention is improved, which is shown in DAN and DCN ($3^{rd}$ row, $5^{th}$ \& $6^{th}$ column of figure-~\ref{fig:res6}). The attention map of the DCN model is more correlated with the ground truth human attention map than the reference model. Thus we observe that with the help of supporting and contrasting exemplar, VQA accuracy is improving. Also, figure- ~\ref{fig:res7} provides attention visualization for DAN and DCN with the help of supporting and contrasting attention.




\subsubsection{Attention visualization of DCN with various Human Attention Maps for VQA} \label{av_2}
\begin{figure*}[!htb]

	\centering
	\includegraphics[width=0.849\textwidth]{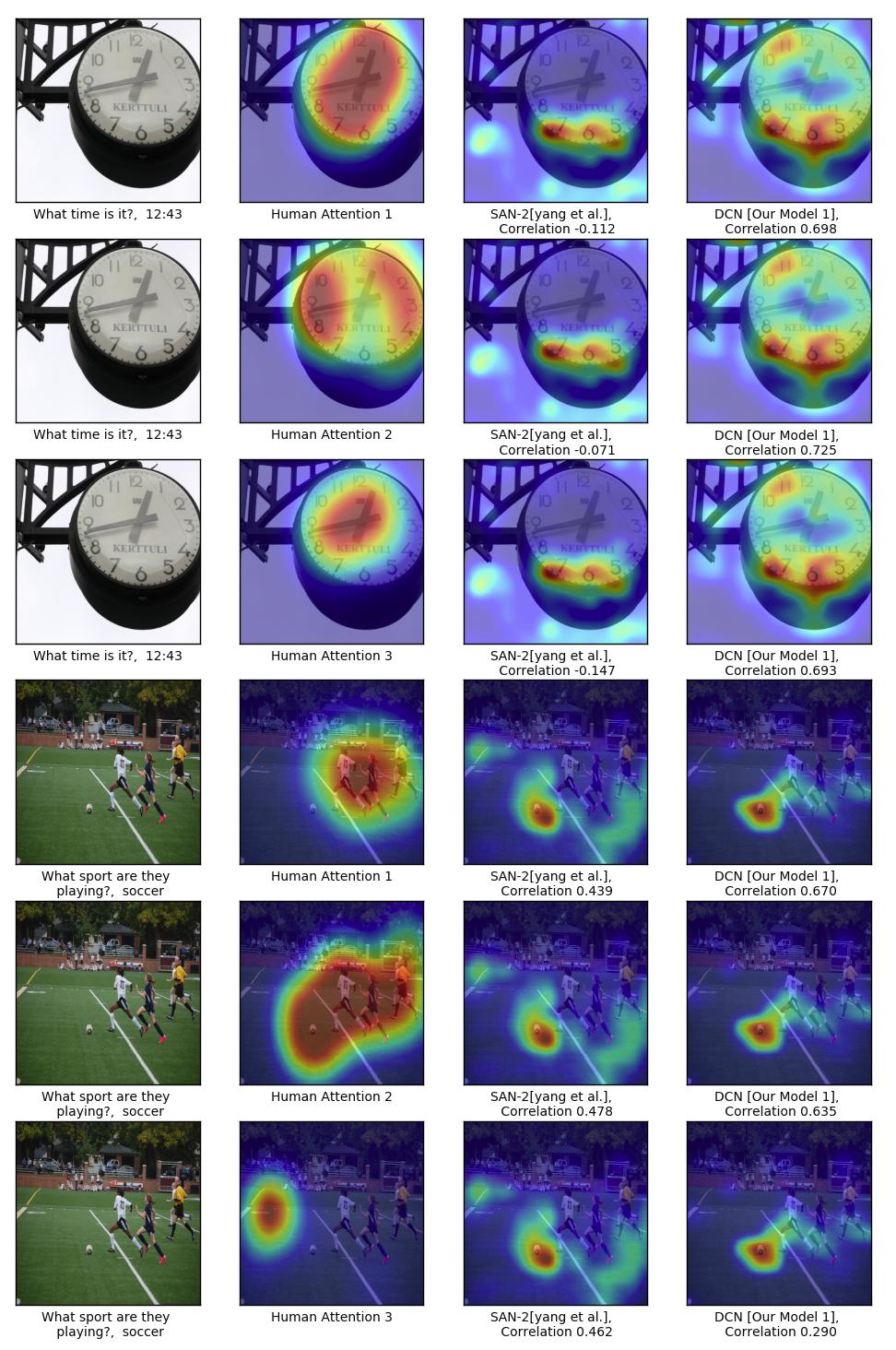}
	\vspace{-1.4em}
	\caption{DCN Attention Map with all ground truth three human attentions. The first row provides results for the human attention map-1 for the HAT dataset\cite{Das_EMNLP2016}. The second row and third row provides results for the human attention map-2 and human attention map-3. Similar results for other examples.}
	\label{fig:res3}
\end{figure*}

\begin{figure*}[!htb]
	\centering
	\includegraphics[width=0.849\textwidth]{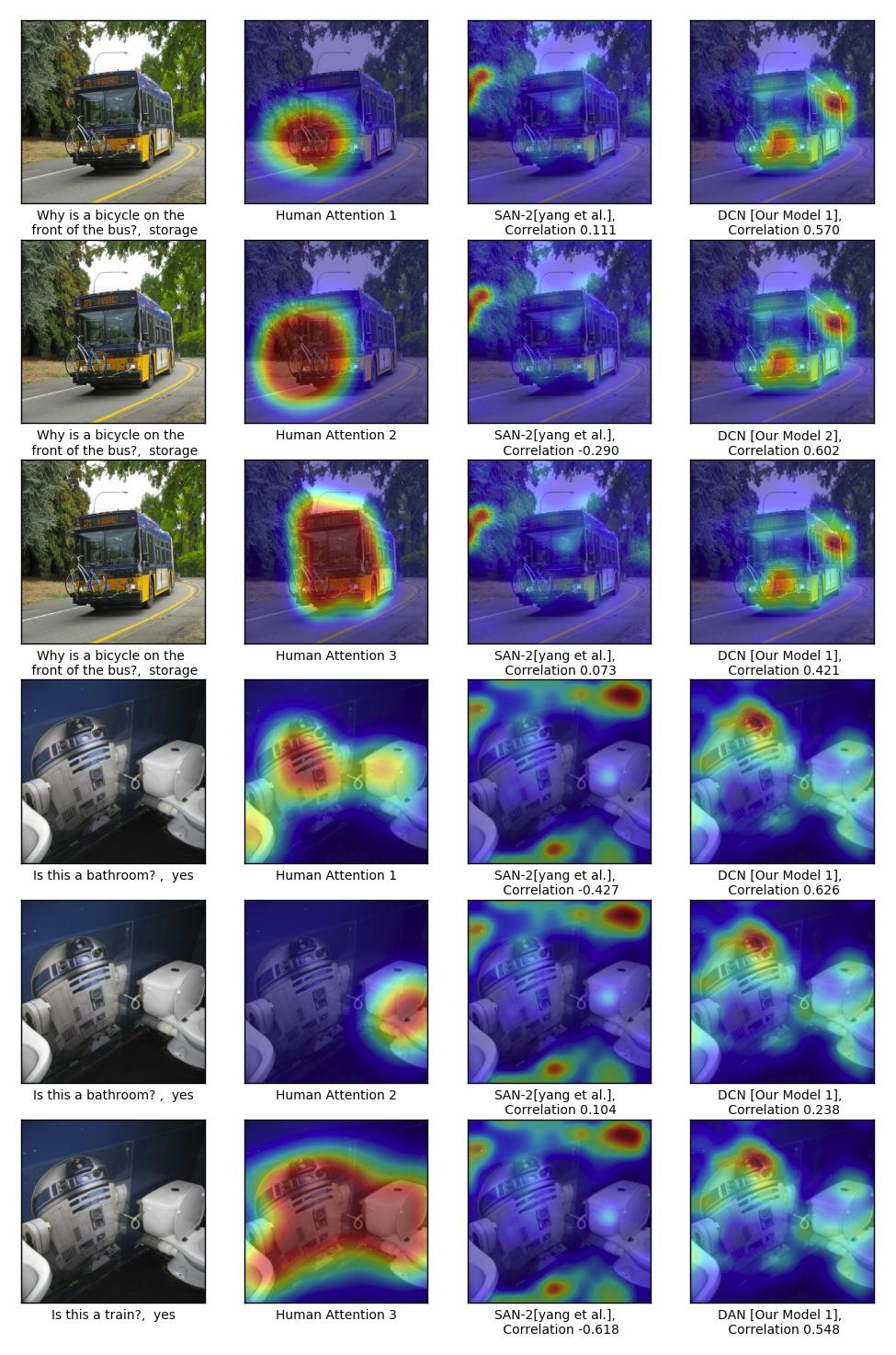}
	\vspace{-1.4em}
	\caption{DCN Attention Map with all ground truth three human attentions. The first row provides results for the human attention map-1 for the HAT dataset\cite{Das_EMNLP2016}. The second row and third row provides results for the human attention map-2 and human attention map-3. Similar results for other examples.}
	\label{fig:res4}
\end{figure*}

We compute rank correlation for all three ground truth human attention maps provide by VQA- HAT\cite{Das_EMNLP2016} Val dataset with our DAN and DCN exemplar model and also visualized attention map with all three ground truth human attention map as shown in figure-~\ref{fig:res3} and ~\ref{fig:res4}. We can evaluate our rank correlation for all three human attention maps and observed that human attention map one is better than attention map-2 and 3 in terms of visualization and rank correlation, as mention in figure-~\ref{fig:res3} and ~\ref{fig:res4}. 

\subsubsection{Attention Visualization of DAN and DCN  for VQA}\label{av_3}
\begin{figure*}[!ht]
	\centering
	\includegraphics[width=0.840\textwidth]{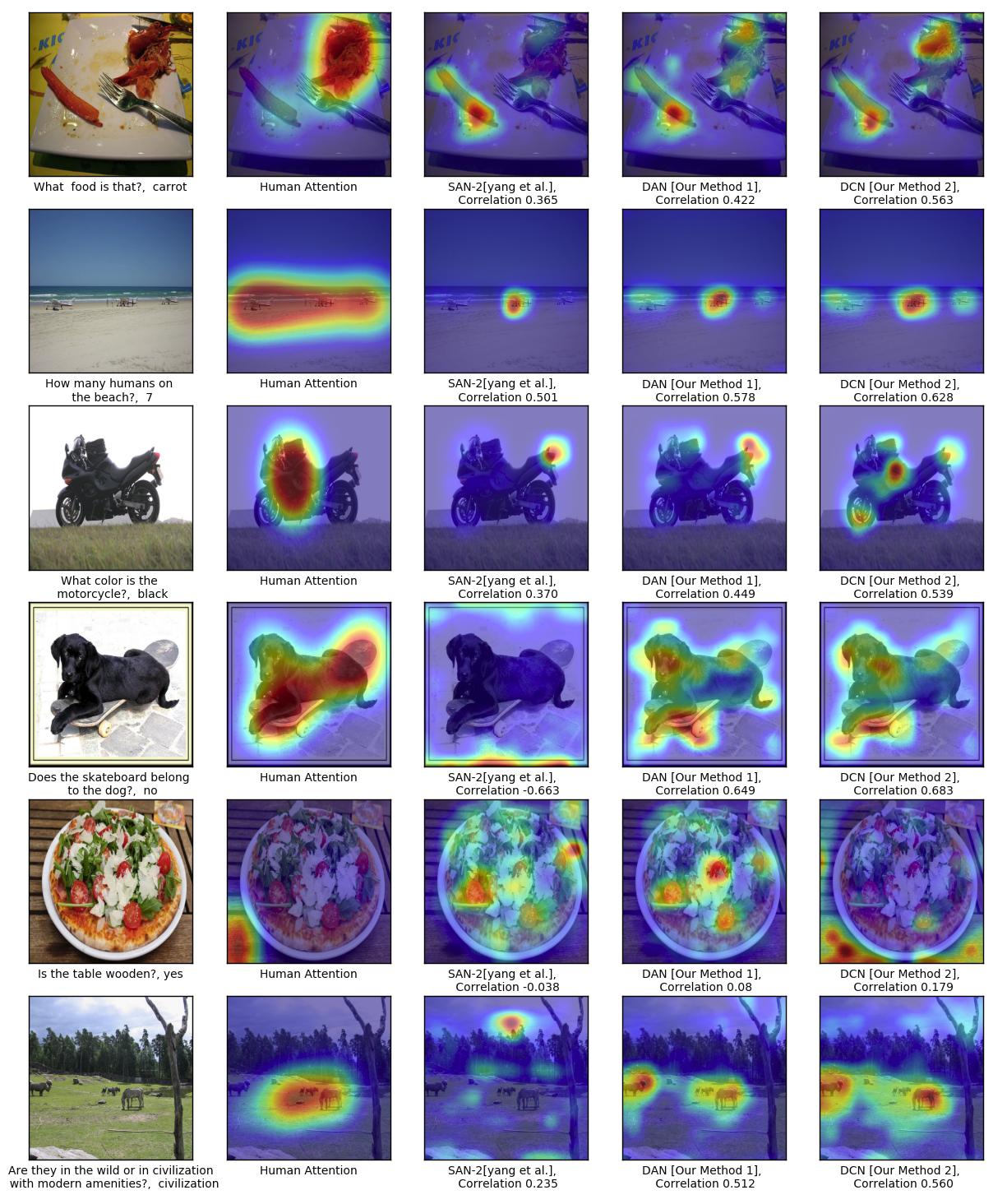}
	\vspace{-1.4em}
	\caption{Attention Result for DAN and DCN. In this figure, the first column indicates target question and corresponding image; second column indicates reference human attention map in HAT dataset, the third column refers to generated attention map for SAN, fourth column refers to the rank correlation of our DAN model and final column refers to rank correlation for our DCN model.}
	\label{fig:res1}
\end{figure*}

\begin{figure*}[!ht]
	\centering
	\includegraphics[width=0.849\textwidth]{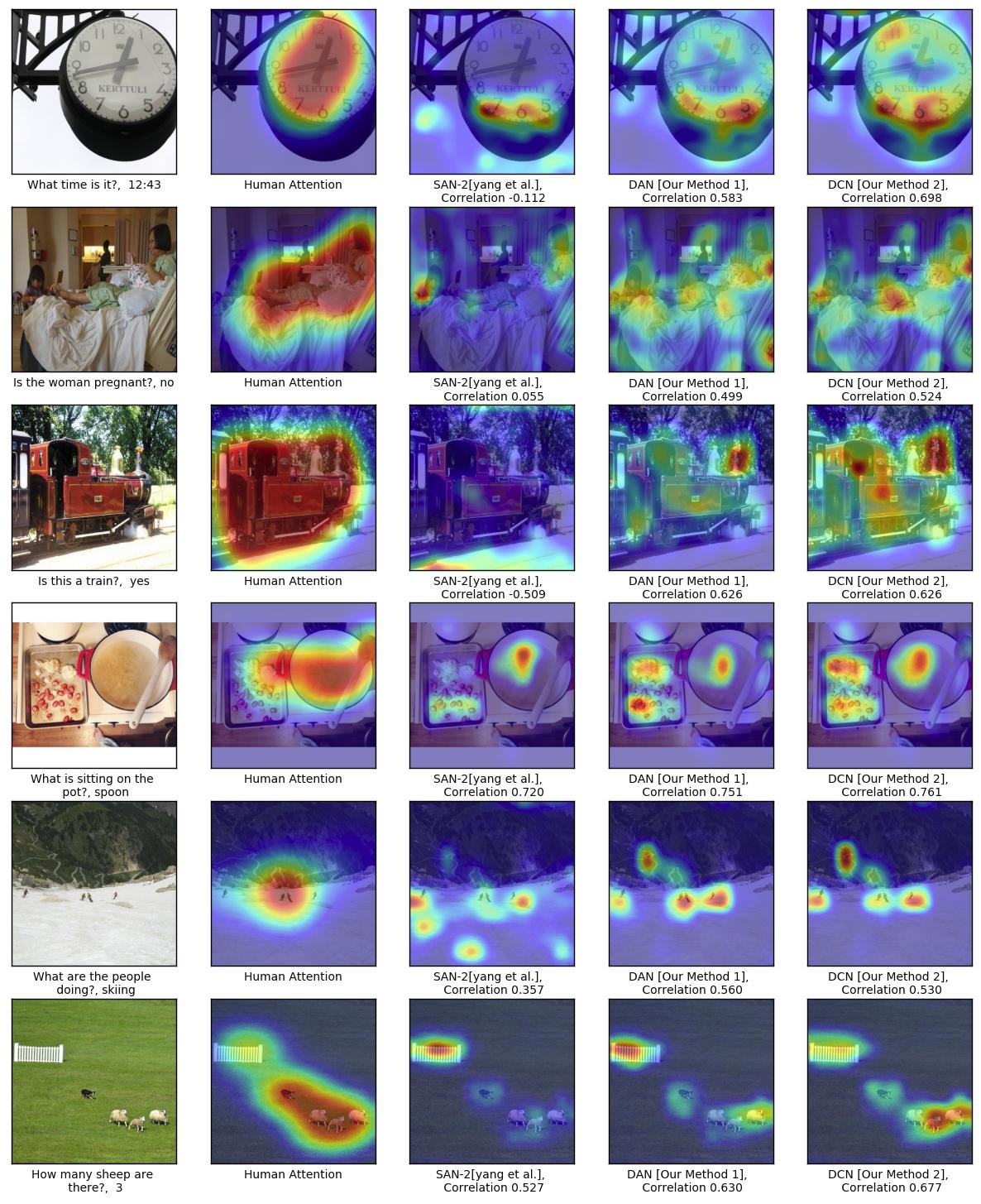}
	\vspace{-1.4em}
	\caption{Attention Result for DAN and DCN. In this figure, the first column indicates target question and corresponding image; second column indicates reference human attention map in HAT dataset, the third column refers to generated attention map for SAN, fourth column refers to the rank correlation of our DAN model and final column refers to rank correlation for our DCN model.}
	\label{fig:res2}
\end{figure*}

We provide the results of the attention visualization in figure-~\ref{fig:res1} and ~\ref{fig:res2}.  As can be observed in figure-~\ref{fig:res1} and ~\ref{fig:res2}, we obtain a significant improvement of rank correlation in attention map by using exemplar model(DCN or DAN) as compared to the SAN method~\cite{Yang_CVPR2016}. We can observe that DAN and DCN have more correlation with human attention.   We observed that DAN and DCN have better rank correlation then SAN attention map.

\subsection{Gradient Computation for Triplet Network} \label{gradient_cal}
	The concept triplet loss is motivated in the context of larger margin nearest neighbor classification\cite{Weinberger_JMLR2009}, which minimizes the distance between the target and supporting feature and maximizes the distance between the target and contrasting feature. $f(x_{i})$ is the embedding feature of $i^{th}$ example of training image $x_{i}$ in n-dimensional euclidean space.
	
	\begin{itemize}
		\item 	$f(s_{i})$ : The embedding of target
		\item	$f(s_{i}^{+})$ :The embedding of supporting exemplar
		\item	$f(s_{i}^{-})$ :The embedding of contrasting exemplar
	\end{itemize}
	
	The objective of triplet loss is to make both supporting features target will have the same identity\cite{Schroff_CVPR2015} \& target, and the contrasting feature will have different identities. Which means it brings all supporting features more close to the target feature than that of contrasting features.
    \begin{equation}
    \begin{split}
    & D(f(s_{i}),f(s_{i}^{+})) +\alpha < D(f(s_{i}),f(s_{i}^{-}))\\
    & \forall{(f(s_{i}),f(s_{i}^{+}),f(s_{i}^{-}))} \in T
    \end{split}
    \end{equation}
    where $D(f(s_{i}),f(s_{j})) = ||f(s_{i})- f(s_{j})||_{2}^{2}$ is defined as the euclidean distance between $f(s_{i}) \& f(s_{j})$. $\alpha$ is the margin between supporting and contrasting feature.The default value of $\alpha$ is 0.2. T is training dataset set, which contain all set of possible triplets. 
    The objective function for  triplet loss is given by 
\begin{dmath}
T(s_i,s_i^+,s_i^-) = \texttt{max}(0, ||f(s_{i})-f(s_{i}^{+})||^{2}_{2} + \alpha - ||f(s_{i})-f(s_{i}^{-})||^{2}_{2})
\end{dmath}
    For simplicity ,the notation are replaced like this , $f(s_{i}) \rightarrow f , f(s_{i}^{+})\rightarrow f^+ , f(s_{i}^{-})\rightarrow f^-$.
    
    Gradient computation of L2 norm is given by  
    \begin{equation}
    \frac{\partial }{\partial x}{||f(x)||^{2}_{2}} = 2*f(x)\frac{\partial }{\partial x}{f(x)}
    \end{equation}

    The gradient of loss w.r.t the "Supporting" input $f^{+}$:
	\[
	\frac{\partial L}{\partial f^+} =\begin{cases}
	\Delta L^+, & \text{if $(\alpha +||f-f^{+}||^{2}_{2} - ||f-f^{-}||^{2}_{2})\ge 0$}\\
	0, & \text{otherwise}.
	\end{cases} 
	\]
	where $\Delta L^+=2*(f-f^+)\frac{\partial (f-f^+)}{\partial f^+}$

	\begin{equation}
	\frac{\partial L}{\partial f^+} =\begin{cases}
	-2(f-f^+), & \text{if $(\alpha +||f-f^{+}||^{2}_{2} - ||f-f^{-}||^{2}_{2})\ge 0$}\\
	0, & \text{otherwise}.
	\end{cases}  
	\end{equation}		
	
	The gradient of loss w.r.t the "Opposing" input $f^{-}$:
	\[
	\frac{\partial L}{\partial f^-} =\begin{cases}
	\Delta L^-, & \text{if $(\alpha +||f-f^{+}||^{2}_{2} - ||f-f^{-}||^{2}_{2})\ge 0$}\\
	0, & \text{otherwise}.
	\end{cases} 
	\]
	where $\Delta L^-=-2*(f-f^-)\frac{\partial (f-f^-)}{\partial f^-}$
	\begin{equation}
	\frac{\partial L}{\partial f^-} =\begin{cases}
	2(f-f^-), & \text{if $(\alpha +||f-f^{+}||^{2}_{2} - ||f-f^{-}||^{2}_{2})\ge 0$}\\
	0, & \text{otherwise}.
	\end{cases}  
	\end{equation}
	
	The gradient of loss w.r.t the "Target" input $f$:
\[
	\frac{\partial L}{\partial f} =\begin{cases}
	\Delta L, & \text{if $(\alpha +||f-f^{+}||^{2}_{2} - ||f-f^{-}||^{2}_{2})\ge 0$}\\
	0, & \text{otherwise}.
	\end{cases} 
\]
where $\Delta L=2*(f-f^+)\frac{\partial (f-f^-)}{\partial f} - 2*(f-f^-)\frac{\partial (f-f^-)}{\partial f}$

	\begin{equation}
	\frac{\partial L}{\partial f} =\begin{cases}
	2(f^--f^+), & \text{if $(\alpha +||f-f^{+}||^{2}_{2} - ||f-f^{-}||^{2}_{2})\ge 0$}\\
	0, & \text{otherwise}.
	\end{cases}  
	\end{equation}

	\subsection{Variant of Triplet Model: Quintuplet Model}
    Unlike the triplet model, In this model, we considered two supporting and two opposing images along with the target image. We have selected supporting and opposing image by clustering. i.e.,    The 2000th nearest neighbor is divided into 20 clusters based on the distance from the target image. That is, the first cluster mean distance is minimum cluster distance from target and 20th  cluster mean distance is the maximum cluster distance from the target. 
	\begin{itemize}
		\item 	$a_i=f(s_{i})$ : The embedding of Target 
		\item	$p_i^{+}=f(s_{i}^{+})$ :The embedding of supporting exemplar from cluster 1 
		\item	$n_i^{-}=f(s_{i}^{-})$ :The embedding of opposing exemplar from cluster 20 
		\item	$p_i^{++}=f(s_{i}^{++})$ :The embedding of supporting exemplar from cluster 2
		\item	$n_i^{--}=f(s_{i}^{--})$ :The embedding of opposing exemplar from cluster 19
	\end{itemize}
	
	The objective of quintuplet is to bring $p_{i}^{+}$(cluster 1) supporting feature more close to target feature than that of $p_{i}^{++}$ (cluster 2) supporting feature than that of $n_{i}^{--}$ (cluster 19) opposing feature than that of $n_{i}^{-}$(cluster 20) opposing feature. 
	
	\begin{equation}
		\begin{split}
	& D(a_{i},p_{i}^{+}) + \alpha_1 < D(a_{i},p_{i}^{++}) 
	+ \alpha_2 < \\
	& D(a_{i},n_{i}^{--})  + \alpha_3< D(a_{i},n_{i}^{-}), \\
    & \forall{(a_{i},p_{i}^{+},p_{i}^{++},n_{i}^{--},n_{i}^{-})} \in T
		\end{split}
	\end{equation}
	where $\alpha_1$,$\alpha_2$,$\alpha_3$ are the margin between $p_{i}^{+} \& p_{i}^{++} $ , $p_{i}^{++} \& n_{i}^{--} $, $n_{i}^{--} \& n_{i}^{-} $ respectively. T is training dataset set, which contain all set of possible quintuplet set. 
	
	Objective function for Quintuplet loss\cite{Huang_CVPR2016} is defined as :
	\begin{equation}
	\min{\sum^{N}_{i=1} (\varepsilon_{i} +\chi_{i}+\phi_{i})+\lambda||\theta||_{2}^{2}}\\
	\end{equation}
	subjected to : \\
	$  max(0,\alpha_1 + D(a_i,p_{i}^{+}) - D(a_i,p_{i}^{++})) \le \varepsilon_{i}\\	
	max(0,\alpha_2 + D(a_i,p_{i}^{++}) - D(a_i,n_{i}^{--})) \le \chi_{i}\\	
	max(0,\alpha_3 + D(a_i,n_{i}^{--})  - D(a_i,n_{i}^{-})) \le \phi_{i}$\\	
	$\forall{i},\varepsilon_{i} \ge 0, \chi_{i} \ge 0,\phi_{i} \ge 0$\\
	
	where $\varepsilon_{i},\phi_{i},\chi_{i}$ are the slack variable  and $\theta$  is the parameter  of attention network and $\lambda$ is a regularizing control  parameter.The value of  $\alpha_1$, $\alpha_2$, $\alpha_3$ are 0.006, 0.2,0.006 set experimentally.

\subsection{VQA Dataset}\label{dataset_hat_vqa}
    We have conducted our experiments on two types of the dataset; first one is VQA dataset, which contains human-annotated question and answer based on images on MS-COCO dataset. The second one is the HAT dataset based on the attention map.  
    \subsubsection{VQA dataset}
    VQA dataset\cite{VQA} is one of the largest dataset for the VQA benchmark so far.  It built on complex images from the ms-coco dataset. VQA dataset contains a total of 204721 images, out of which 82783 images for training, 40504 images for validation, and 81434 images for testing. Each image in the MS-COCO dataset\cite{Lin_ECCV2014} is associated with 3 questions, and each question has 10 possible answers. This dataset is annotated by different people. So there are  248349 QA pairs for training, 121512 QA pairs for validating, and 244302 QA pairs for testing. 
    We use the top 1000 most frequently output as our possible answer set, as is commonly used. This covers 82.67\%  of the train+val answer.

    \subsubsection{VQA-HAT(Human Attention) dataset}
    We used VQA-HAT dataset\cite{Das_EMNLP2016}, which is developed based on the de-blurring task to answer visual questions. 
    This dataset contains a human attention map for the training set of 58475 examples out of 248349 VQA training set. It contains 1374 validation examples out of 121512 examples of question image pair in the VQA validation set.

	\begin{algorithm*}
		
		\caption{Rank Correlation Procedure}\label{Correlation}
		\begin{algorithmic}[1]
			\Procedure{:}{Initialization}
			\State $P_{HAM}$: Probability distribution of Human Attention Map
			\State $P_{DAN}$	: Probability distribution of Differential Attention 
			\BState \emph{\textbf{Rank}}:
			\State Compute Rank of $P_{HAM}$ : $R_{HAM}$
			\State Compute Rank of $P_{DAN}$ : $R_{DAN}$ 
			
			\BState \emph{\textbf{Rank Difference }}:
			
			\State Compute difference in rank between $R_{HAM}$ \& $R_{DAN}$ : $Rank_{Diff}$
			\State Compute square of rank difference $Rank_{Diff}$ :$S_{Rank\_Diff}$

			\BState \emph{\textbf{Rank Correlation}}:
			\State Compute Dimension of $P_{DAN}$ :N
			\State Compute Rank Correlation using :
			\[R_{Cor}= 1- {\frac{6*S_{Rank\_Diff}}{N^3 - N}}\]
			
			\EndProcedure
		\end{algorithmic}
	\end{algorithm*}

\subsection{Training and Evaluation methods}\label{train_eval}
In this section we provide details on training setup and model configuration for VQA as in section-\ref{train_vqa}, details evaluating VQA accuracy on VQA dataset as in section-\ref{vqa_dataset} and Evaluating Rank Correlation on HAT dataset hat dataset on section-\ref{hat_dataset}.

\subsubsection{ Training Setup and Model Configuration for VQA}\label{train_vqa}
    We have extracted the CONV-5 image feature of the pre-trained VGG-19 CNN model for the LSTM + Q+ I+ Attention baseline model. Since submission, we have updated our model and used the CONV-5 image feature of the pre-trained Resnet-152 CNN model for MCB to get state of the art result. We trained the differential attention model using joint loss in an end-to-end manner. We have used RMSPROP  optimizer to update the model parameter and configured hyper-parameter values to be as follows: {learning rate =0.0004 , batch size = 200, alpha = 0.99 and epsilon=1e-8} to train the classification network . In order to train a triplet model, we have used RMSPROP to  optimize the triplet model model parameter and configure hyper-parameter values to be: {learning rate =0.001 , batch size = 200, alpha = 0.9 and epsilon=1e-8}. We have used learning rate decay to decrease the learning rate on every epoch by a factor given by:
    \[Decay\_factor=exp\left(\frac{log(0.1)}{a*b} \right)\] where value of  a=1500 and b=1250 is set empirically. The selection of training controlling factor($\nu$) has a major role during training. If $\nu$=1 means updating the triplet and classification network parameter at the same rate. If $\nu$ $\gg$ 1 means updating the triplet net more frequently as compare to the classification net. Since triplet loss decreases much lower then classification loss, we fixed the value of $\nu$ $\gg$ 1 that is a fixed value of $\nu$=10.
    
    \subsubsection{Evaluating VQA Accuracy on VQA dataset}\label{vqa_dataset}
    VQA dataset contain 3 type of answer: yes/no, number  and other. 
    The evaluation is carried out using two test splits,i.e test-dev and test-standard. The question in corresponding test split are answered using two ways: Open-Ended\cite{VQA} and Multiple-choice. Open-Ended task should generate a natural language answer in form of single word or phrase. For each question there are 10 candidate answer provided with their respective confidence level. Our module generates a single word answer on the open ended task. This answer can be evaluated using accuracy metric provide by Antol {\it et al.}\cite{VQA} as follows. 
    \begin{equation} 
    Acc=\frac{1}{N}\sum_{i=1}^{N}{min\big(\frac{\sum_{t \in T^{i} }{\mathbb {I}[a_i=t]}}{3},1 \big)}
    \end{equation}
    
    Where $a_i$ the predicted answer and t is the annotated answer in the target answer set $T^{i}$  of the $i^{th}$ example and $\mathbb {I}[.], $ is the indicator function.
    The predicted answer $a_i$ is correct if at least 3 annotators agree on the predicted answer. If the predicted answer is not correct, then the accuracy score depends on the number of annotators that agree on the answer. Before checking the accuracy, we need to convert the predicted answer to lowercase, the number to digits and punctuation \& article to be removed.          

    \subsubsection{Evaluating Rank Correlation on HAT dataset}\label{hat_dataset}
    We used rank correlation technique to evaluate\cite{Das_EMNLP2016} the correlation between human attention map and DAN attention probability. Here we scale down the human attention map to 14x14 in order to make the same size as DAN attention probability. We then compute rank correlation using the following steps. 
    Rank correlation technique is used to obtain the degree of association between the data.  The value of rank correlation\cite{Mcdonald_handbook2009} lies between +1 to -1. When $R_{Cor}$ is close to 1, it indicates a positive correlation between them, When $R_{Cor}$ is close to -1, it indicates a negative correlation between them, and when $R_{Cor}$ is close to 0, it indicates No correlation between them. A higher value of rank correlation is better. Rank correlation of predicted attention map with human attention map can be obtained using the following algorithm-\ref{Correlation}.

\begin{table*}[ht]
\centering
\caption{\label{score_tab_3_vqg_full}Full State-of-the-Art comparison on VQG-COCO Dataset. The first block consists of state-of-the-art results, the second block refers to the baselines mentioned in the State-of-the-art section of the main paper, and the third block provides the results for the best method for different ablations of our method.}
\begin{tabular}{|l|lcccccc|}
\hline \bf Context &  \bf BLEU1 & \bf BLEU2  & \bf BLEU3  & \bf BLEU4& \bf METEOR & \bf ROUGE & \bf CIDEr \\ \hline 

Natural\cite{mostafazadeh2016generating} & 19.2  & -&- & -  & 19.7  &- & -  \\
Creative\cite{jain2017creativity} & {35.6}  & -& -& -&   19.9 & - & - \\ \hline
Image Only &  20.8  & 14.1&8.5 &5.2 &     8.6  & 22.6 & 18.8\\
Caption Only & 21.1  & 14.2& 8.6&5.4 &   8.5  & 25.9 & 22.3\\\hline
Tag-Hadamard & {24.4}  &15.1 & 9.5& 6.3&{10.8}& {24.3} & {55.0}\\ 
PlaceCNN-Joint & {25.7}  & 15.7& 9.9& 6.5  &{10.8}&{24.5} &\textbf{56.1} \\
Diff.Image-Joint& 30.4  & 20.1& 14.3& 8.3 & {11.7} & 26.3 & {38.8}\\
MDN-Joint (\bf Ours)& \textbf{36.0}  & 24.9&16.8 & 10.4&\textbf{23.4}&\textbf{41.8}& 50.7\\\hline
Humans\cite{mostafazadeh2016generating} & \textbf{86.0} &- &- &- &\textbf{60.8}&\textbf{-}& -\\\hline
\end{tabular}

\end{table*}
\section{Experiments on VQG Task}\label{exp_vqg}
\begin{figure*}[ht]
	\centering
	\includegraphics[width=0.9 \textwidth]{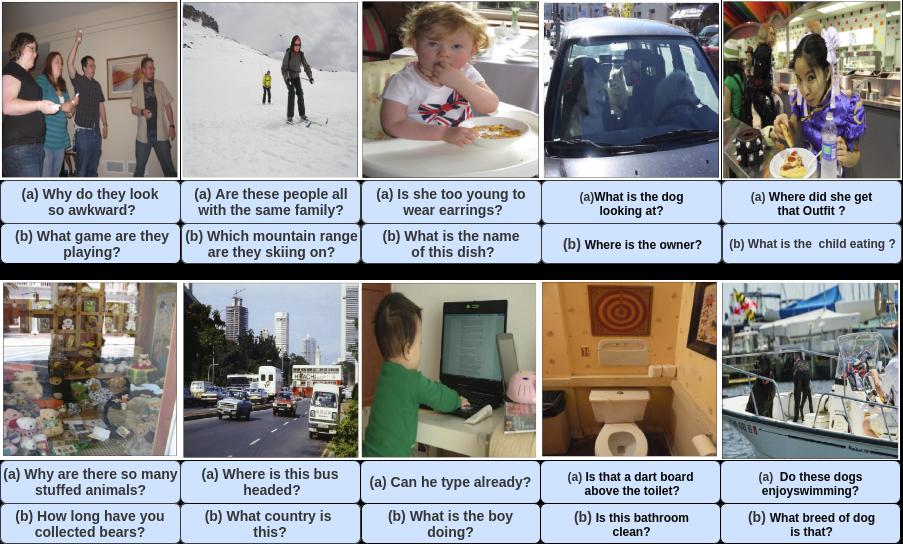}
	\vspace{-0.2cm}
	\caption{These are some examples from the VQG-COCO dataset, which provide a comparison between our generated questions and human-annotated questions. (a) is the human-annotated question for all the images.}
	\label{fig:natural2}
\end{figure*}

We evaluate our proposed MDN method in the following ways: First, we further compare our network with state-of-the-art methods for the VQG-COCO dataset with all scores as in table-\ref{score_tab_3_vqg_full}.
We perform a user study to gauge human opinion on the naturalness of the generated question and analyze the word statistics in Figure~\ref{tbl:sunburst}. This is an important test as humans are the best deciders of naturalness, as mentioned in section-\ref{percep_real}. We further consider the statistical significance for the various ablations as well as the state-of-the-art models in section-\ref{ssa}. The quantitative evaluation is conducted using standard metrics like BLEU~\cite{Papineni_ACL2002}, METEOR~\cite{Banerjee_ACL2005}, ROUGE~\cite{Lin_ACL2004}, CIDEr~\cite{Vedantam_CVPR2015} as shown in figure-\ref{fig:natural2} and \ref{fig:natural2}. Although these metrics have not been shown to correlate with the `naturalness' of the question, these still provide a reasonable quantitative measure for comparison.  Details ablation analysis for the multimodal differential network is described in section-\ref{disc}.
We observe that the proposed MDN provides improved embeddings to the decoder. We believe that these embeddings capture instance-specific differential information that helps in guiding the question generation.

\begin{figure*}[ht]
	\centering
	\includegraphics[width=0.9 \textwidth]{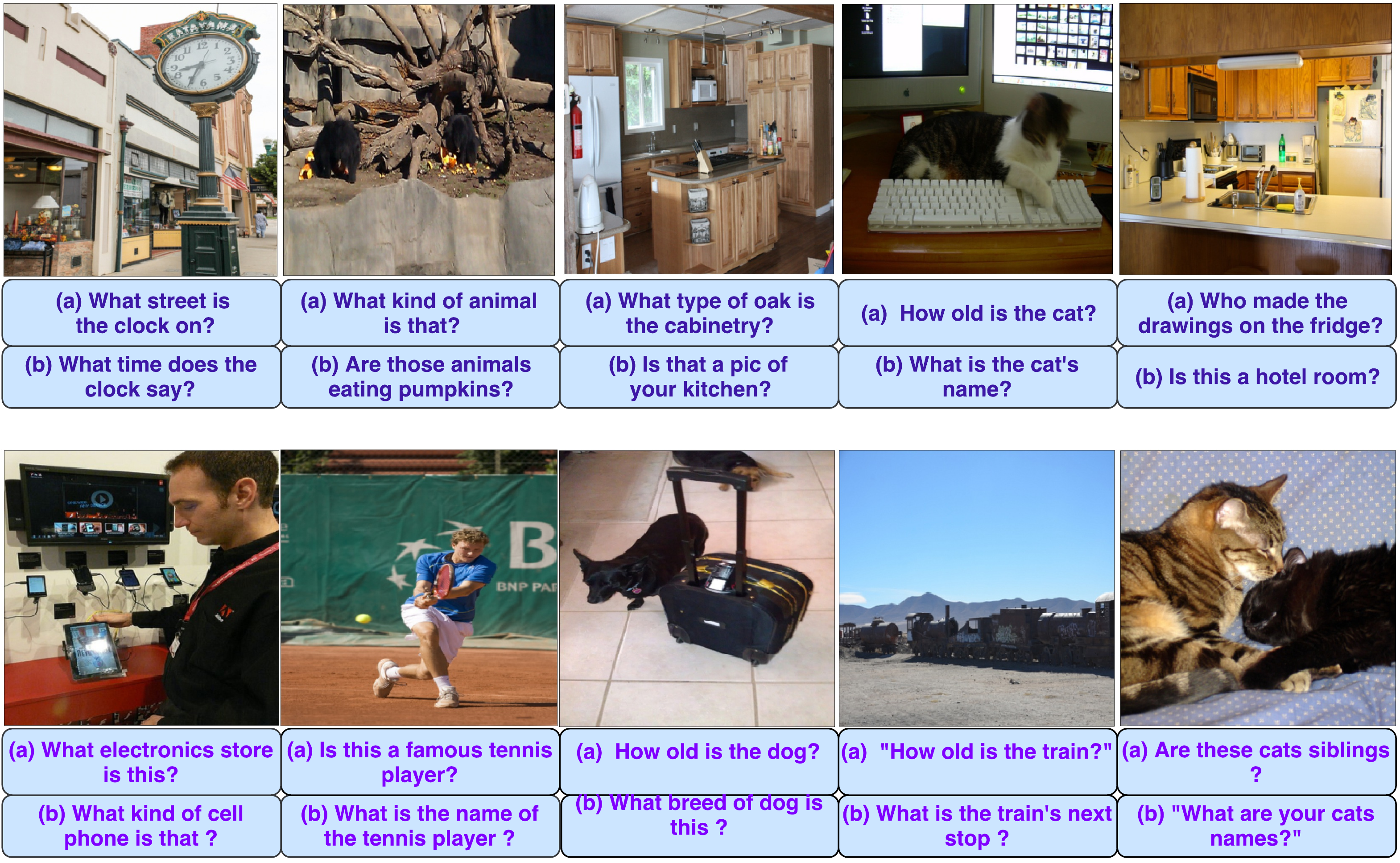}
	\caption{These are some more examples from the VQG-COCO dataset, which provide a comparison between the questions generated by our model and human-annotated questions. (b) is the human-annotated question for the first row-fourth column, \& fifth column image, and (a) for the rest of images.}
	\label{fig:natural3}
\end{figure*}

\subsection{Statistical Significance Analysis}\label{ssa}
\begin{figure}[ht]
	\centering
	\includegraphics[width=0.45\textwidth]{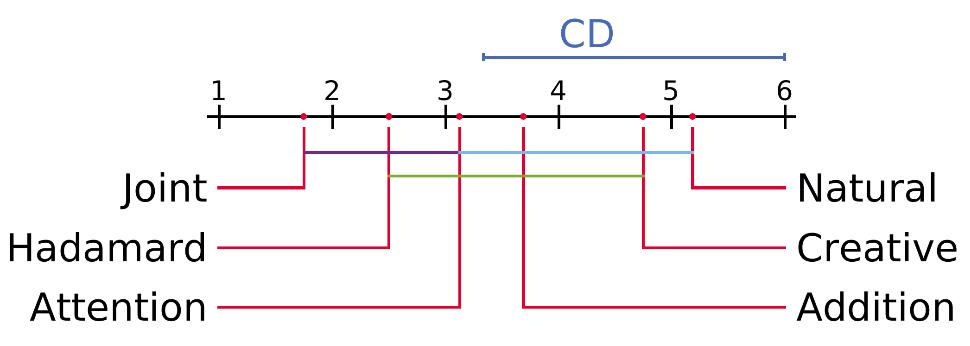}
	\vspace{-0.35cm}
	\caption{The mean rank of all the models on the basis of the METEOR score is plotted on the x-axis. Here Joint refers to our MDN-Joint model, and others are the different variations described in section-3.3.2 and Natural~\cite{mostafazadeh2016generating}, Creative~\cite{jain2017creativity}. The colored lines between the two models represent that these models are not significantly different from each other.}
	\label{fig:result_1_B}
\end{figure}

\begin{figure}[ht]
	\centering
	\includegraphics[width=0.5\textwidth]{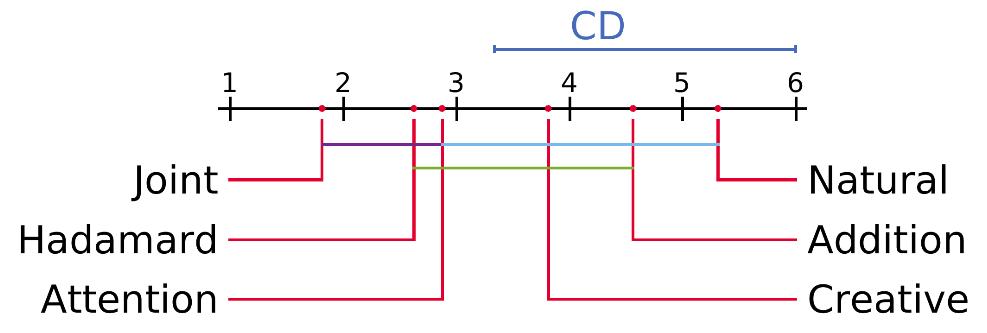}
	\caption{The mean rank of all the models based on the BLEU score is plotted on the x-axis. Here Joint refers to our MDN-Joint model, and others are the different variations of our model and Natural-\cite{mostafazadeh2016generating}, Creative-\cite{jain2017creativity}. Also, the colored lines between the two models represent that those models are not significantly different from each other. }
	\label{fig:result_1_A}
\end{figure}

We have analysed Statistical Significance~\cite{Demvsar_JMLR2006} of our MDN model for VQG for different variations of the mixture module mentioned in section-3.3.2 and also against the state-of-the-art methods. 
The Critical Difference (CD) for Nemenyi~\cite{Fivser_PLOS2016} test depends upon the given $\alpha$ (confidence level, which is 0.05 in our case) for average ranks and N (number of tested datasets). If the difference in the rank of the two methods lies within CD, then they are not significantly different and vice-versa. Figure~\ref{fig:result_1_B} visualizes the post hoc analysis using the CD diagram. From the figure, it is clear that MDN-Joint works best and is statistically significantly different from the state-of-the-art methods. Also, we analyze our model based on the BLEU score, as shown in Figure~\ref{fig:result_1_A}. We got a similar type of behavior in both BLEU and METEOR score, as shown in figure-~\ref{fig:result_1_A} and ~\ref{fig:result_1_B}. 

\begin{figure}[ht]
	\centering
	\includegraphics[width=0.5\textwidth]{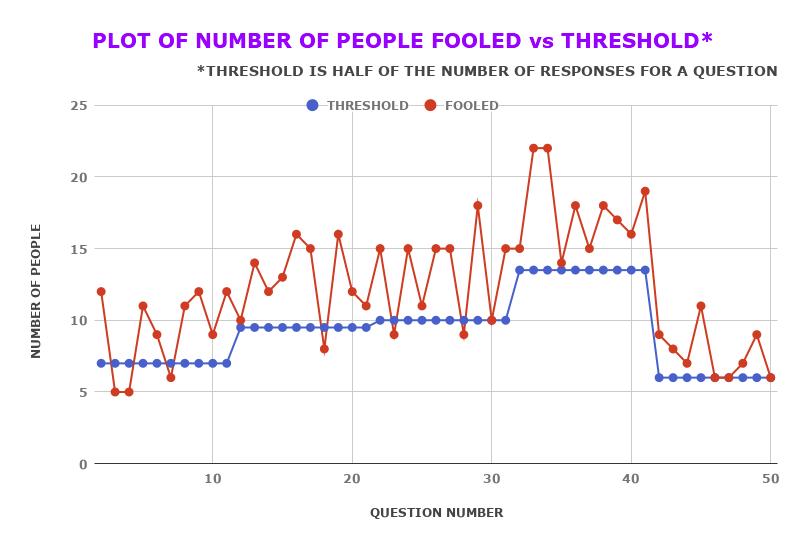}
	\vspace{-0.89cm}
	\caption{Perceptual Realism Plot for the human survey. Here every question has a different number of responses, and hence the threshold, which is half of the total responses for each question is varying. This plot is only for 50 of the 100 questions involved in the survey. See section~\ref{percep_real} for more details.}
	\label{fig:result_2_A}
\end{figure}

\begin{figure}[ht]
	\centering
	\includegraphics[width=0.4\textwidth]{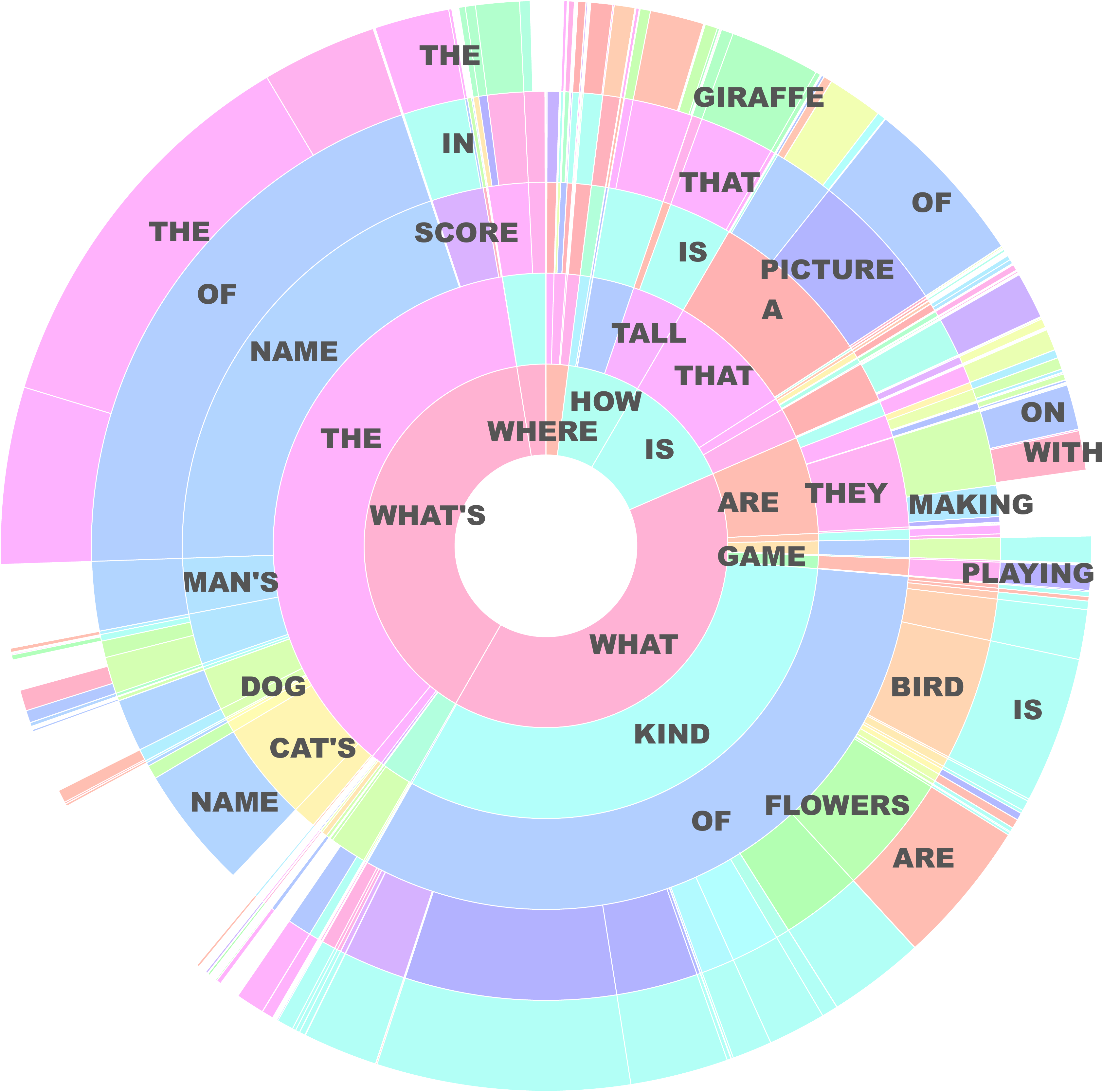}
	\caption{Sunburst plot for VQG-COCO: The $i^{th}$ ring captures the frequency distribution over words for the $i^{th}$ word of the generated question. The angle subtended at the center is proportional to the frequency of the word. While some words have high frequency, the outer rings illustrate a fine blend of words. We have restricted the plot to 5 rings for easy readability. Best viewed in color.}
	\label{tbl:sunburst}
\end{figure}


\subsection{Perceptual Realism}\label{percep_real}
A human is the best judge of the naturalness of any question; We evaluated our proposed MDN method using a `Naturalness' Turing test~\cite{Zhang_ECCV2016} on 175 people. 
People were shown an image with 2 questions just as in figure~\ref{fig:natural2} and were asked to rate the naturalness of both the questions on a scale of 1 to 5  where 1 means `Least Natural' and 5 is the `Most Natural'. We provided 175 people with 100 such images from the VQG-COCO validation dataset, which has 1250 images. 
Figure~\ref{fig:result_2_A} indicates the number of people who were fooled (rated the generated question more or equal to the ground truth question). For the 100 images, on an average 59.7\% people were fooled in this experiment, and this shows that our model is able to generate natural questions. 

\section{More Ablation Analysis for VQG task}\label{disc}

\subsection{How are exemplars improving Embedding}
In the Multimodel differential network, we use exemplars and train them using a triplet loss. It is known that using a triplet network; we can learn a representation that accentuates how the image is closer to a supporting exemplar as against the opposing exemplar~\cite{Hoffer_Springer2015, Frome_ICCV2007}. The Joint embedding is obtained between the image and language representations. Therefore the improved representation helps in obtaining an improved context vector. Further, we show that this also results in improving VQG. 

\subsection{Are exemplars required?}
We had similar concerns and validated this point by using random exemplars for the nearest neighbor for MDN. (k=R in table~\ref{score_tab_11}) In this case, the method is similar to the baseline. This suggests that with random exemplar, the model learns to ignore the cue. 

\subsection{Are captions necessary for our method?}
 This is not actually necessary. In our method, we have used an existing image captioning method~\cite{Karpathy_CVPR2015} to generate captions for images that did not have them. For the VQG dataset, captions were available, and we have used that, but, for the VQA dataset, captions were not available, and we have generated captions while training. We provide detailed evidence with respect to caption-question pairs to ensure that we are generating novel questions. While the caption generates scene descriptions, our proposed method generates semantically meaningful and novel questions.
 Examples for Figure 1 of the main paper:
First Image:- Caption- A young man skateboarding around little cones. Our Question- Is this a skateboard competition?
Second Image:- Caption- A small child is standing on a pair of skis.
Our Question:- How old is that little girl?
\subsection{Sampling Exemplar: KNN vs ITML}
Our method is aimed at using efficient exemplar-based retrieval techniques. 
We have experimented with various exemplar methods, such as  ITML \cite{davis_ACM2007} based metric learning for image features and KNN based approaches. We observed a KNN based approach (K-D tree) with the Euclidean metric is an efficient method for finding exemplars. Also, we observed that ITML is computationally expensive and also depends on the training procedure. The table provides the experimental result for Differential Image Network variant with k (number of exemplars) = 2  and Hadamard method:	

 \begin{table}[h!]
\centering
\caption{\label{score_tab_13}VQG-COCO-dataset, Analysis of different methods of finding Exemplars for Hadamard model. ITML vs. KNN based methods. We see that both give more or less similar results, but since ITML is computationally expensive and the dataset size is also small, it is not that efficient for our use. All these experiments are for the differential image network for K=2 only.}
\begin{tabular}{|l|c|cccc|}
\hline \bf Meth & \bf Exemplar & \bf BLEU-1 & \bf Meteor & \bf Rouge  & \bf CIDer \\ \hline
KNN &IE(K=2)& 23.2 &8.9 & \textbf{27.8} & 22.1\\ 
ITML &IE(K=2)& 22.7 &9.3 & 24.5 & 22.1\\ \hline
\end{tabular}

\end{table}

\subsection{Question Generation approaches: Sampling vs Argmax}
 We obtained the decoding using standard practice followed in the literature~\cite{Sutskever_NIPS2014}. This method selects the argmax sentence. Also, we evaluated our method by sampling from the probability distributions and provided the results for our proposed MDN-Joint method for VQG dataset as follows:
 \begin{table}[h!]
\centering
\caption{\label{score_tab_12}VQG-COCO-dataset, Analysis of question generation approaches:sampling vs Argmax in MDN-Joint model for K=5 only. We see that Argmax clearly outperforms the sampling method.}
\begin{tabular}{|l|cccc|}
\hline \bf Meth &\bf BLEU-1 & \bf Meteor & \bf Rouge  & \bf CIDer \\ \hline
Sampling &  17.9 &11.5 & 20.6 & 22.1\\ 
Argmax &  36.0 &23.4 & 41.8 & 50.7\\ \hline
\end{tabular}

\end{table}


\subsection{Network Ablation Analysis}

\label{sec-4}
 While we advocate the use of the multimodal differential network (MDN) for generating embeddings that can be used by the decoder for generating questions, we also evaluate several variants of this architecture, namely (a) Differential Image Network, (b) Tag net and  (c) Place net. These are described in detail as follows:


\begin{table}[h!]
\caption{\label{score_tab_7}  Analysis of different Tags for VQG-COCO-dataset. We analyse noun tag (Tag-n), verb tag (Tag-v), and question tag (Tag-wh) for different fusion methods, namely joint, attention, Hadamard, and addition based fusion.}
\centering
\begin{tabular}{|l|l|cccc|}
\hline \bf Context & \bf Meth & \bf BLEU-1 & \bf Meteor & \bf Rouge & \bf CIDer \\ \hline 
Image & - & 23.2  &  8.6  & 25.6 & 18.8\\
Caption  &- & 23.5 & 8.6  & 25.9 & 24.3\\\hline
Tag-n & JM &22.2 &10.5 & 22.8 &50.1 \\
Tag-n & AtM & 22.4 &8.6 & 22.5 & 20.8\\
Tag-n & HM &\bf24.8&\bf10.6& 24.4 & \bf53.2\\
Tag-n & AM &24.4 &10.6& 23.9 & 49.4\\ \hline

Tag-v & JM & 23.9 &10.5  &24.1 &\bf52.9 \\
Tag-v & AtM & 22.2 &8.6 &22.4 &20.9\\
Tag-v & HM &\bf24.5 &\bf10.7&24.2 &\bf52.3 \\
Tag-v & AM & 24.6 &10.6 &24.1 &49.0\\ \hline

Tag-wh & JM &22.4  &10.5 &22.5 &48.6\\
Tag-wh & AtM &22.2  &8.6 &22.4 &20.9\\
Tag-wh & HM & \bf24.6 &\bf10.8& 24.3 & \textbf{55.0}\\
Tag-wh & AM &24.0  &10.4 &23.7 &47.8\\ \hline

\end{tabular}

\end{table}

\begin{table}[h!]
\centering
\caption{\label{score_tab_9}Combination of 3 tags of each category for hadamard mixture model namely addition, concatenation, multiplication and 1d-convolution}
\begin{tabular}{|l|l|c  c c|}
\hline \bf Context & \bf BLEU-1 & \bf Meteor & \bf Rouge & \bf CIDer \\ \hline

Tag-n3-add & 22.4 &9.1& 22.2 &26.7\\
Tag-n3-con  &\textbf{24.8}&10.6& 24.4 & \bf53.2\\
Tag-n3-joint& 22.1&8.9&21.7 & 24.6\\
Tag-n3-conv  &24.1&10.3&24.0&47.9\\\hline

Tag-v3-add  &24.1 &10.2& 23.9  &46.7 \\
Tag-v3-con  &24.5 &10.7&24.2 &\bf52.3 \\
Tag-v3-joint  & 22.5&9.1&22.1& 25.6\\
Tag-v3-conv  & 23.2&9.0&24.2 &38.0 \\\hline

Tag-q3-add  & 24.5 &10.5& 24.4 &{51.4}\\
Tag-q3-con  & 24.6 &\textbf{10.8}& 24.3 & \textbf{55.0}\\
Tag-q3-joint  & 22.1 &9.0& 22.0 &25.9\\
Tag-q3-conv  &24.3 &10.4&24.0 &48.6\\\hline
\end{tabular}

\end{table}

\subsubsection{Analysis of Context: Tags}
\label{sec:context_analysis_tag}

Tag is a language-based context. These tags are extracted from the caption, except question-tags, which is fixed as the 7 'Wh words' (What, Why, Where, Who, When, Which, and How). We have experimented with Noun tag, Verb tag, and 'Wh-word' tag, as shown in tables. Also, we have experimented in each tag by varying the number of tags from 1 to 7. We combined different tags using 1D-convolution, concatenation, and the addition of all the tags and observed that the concatenation mechanism gives better results.  

As we can see in table~\ref{score_tab_7} that taking Nouns, Verbs, and Wh-Words as context, we achieve significant improvement in the BLEU, METEOR, and CIDEr scores from the basic models which only takes the image and the caption respectively.
Taking Nouns generated from the captions and questions of the corresponding training example as context, we achieve an increase of 1.6\% in Bleu Score and 2\% in METEOR and 34.4\% in CIDEr Score from the basic Image model. Similarly, taking Verbs as context gives us an increase of 1.3\% in Bleu Score and 2.1\% in METEOR and 33.5\% in CIDEr Score from the basic Image model. And the best result comes when we take 3 Wh-Words as context and apply the Hadamard Model with concatenating the 3 WH-words. \\
Also, in Table \ref{score_tab_9}, we have shown the results when we take more than one word as context. Here we show that for 3 words i.e., 3 nouns, 3 verbs, and 3 Wh-words, the Concatenation model performs the best. In this table, the convolution model is using 1D convolution to combine the tags, and the joint model combines all the tags.
\label{sec:context_analysis}

\subsubsection{Analysis of Context: Exemplars }
\label{sec:model_analysis}
In Multimodel Differential Network and Differential Image Network, we use exemplar images(target, supporting, and opposing image) to obtain the differential context. We have performed the experiment based on the single exemplar(K=1), which is one supporting and one opposing image along with the target image, based on two exemplars (K=2), i.e., two supporting and two opposing images along with single target image. similarly, we have experimented K=3 and K=4 as shown in table-~\ref{score_tab_11}.

\begin{table}[ht]
\caption{\label{score_tab_11}VQG-COCO-dataset, Analysis of different number of Exemplars for addition model, hadamard model and joint model, R is random exemplar. All these experiment are for the differential image network. k=5 performs the best and hence we use this value for the results in main paper.}
\centering
\begin{tabular}{|l|c|cccc|}
\hline \bf Meth & \bf Exemplar & \bf BLEU-1 & \bf Meteor & \bf Rouge  & \bf CIDer \\ \hline
AM &IE(K=1)& 21.8 &7.6 & 22.8 & 22.0\\ 
AM &IE(K=2)& 22.4 &8.3 & 23.4 & 16.0\\ 
AM &IE(K=3)& 22.1 &8.8 & 24.7 & 24.1\\ 
AM &IE(K=4)& 23.7 &9.5 &\textbf{ 25.9} & 25.2\\ 
AM &IE(K=5)&\textbf{24.4}  & \textbf{11.7} & 25.0 & {27.8}\\
AM &IE(K=R)& 18.8 & 6.4  & 20.0 & 20.1\\\hline 
HM &IE(K=1)& 23.6 &7.2 & 25.3 & 21.0\\ 
HM &IE(K=2)& 23.2 &8.9 & \textbf{27.8} & 22.1\\ 
HM &IE(K=3)& 24.8 &9.8 & 27.9 & 28.5\\ 
HM &IE(K=4) & 27.7 &9.4 & 26.1 & \textbf{33.8}\\ 
HM &IE(K=5)& \textbf{28.3} &\textbf{10.2}  & 26.6 & 31.5\\
HM &IE(K=R)& 20.1 & 7.7  & 20.1 & 20.5\\\hline 
JM &IE(K=1)& 20.1 &7.9 & 21.8 & 20.9\\ 
JM &IE(K=2)& 22.6 &8.5 & 22.4 & 28.2\\\
JM &IE(K=3)  & 24.0 &9.2 & 24.4 & 29.5\\
JM &IE(K=4)& 28.7 &10.2 & 24.4 & 32.8\\ 
JM &IE(K=5) &\textbf{30.4}  & \textbf{11.7} & \textbf{26.3} & {38.8}\\ 
JM &IE(K=R)& 21.8 & 7.4  & 22.1 & 22.5\\\hline 
\end{tabular}

\end{table}

\section{Experimental Setup on VQG}\label{sec:exp}
It consists of details of datasets upon which we evaluate our method, evaluation methods, and training and model configuration. Algorithm-\ref{MC-BMN} provides pseudo-code for the multimodal differential network.   
\subsection{Dataset for VQG task}
    We conduct our experiments on two types of the dataset; first one is VQA dataset, which contains human-annotated question and answer based on images on MS-COCO dataset. The second one is the VQG dataset based on the naturalness of the question.  
    \subsubsection{VQA dataset}
    VQA dataset\cite{VQA} is one of the largest datasets for the VQA benchmark so far.  It built on complex images from the ms-coco dataset. VQA dataset, each image in the MS-COCO dataset\cite{Lin_ECCV2014} is associated with 3 questions, and each question has 10 possible answers. This dataset is annotated by different people. So there are  248349 QA pairs for training, 121512 QA pairs for validating, and 244302 QA pairs for testing. 
    We use the top 1000 most frequently output as our possible answer set, as is commonly used. This covers 82.67\%  of the train+val answer.

\subsubsection{VQG Dataset}
We conduct our experiments on the Visual Question Generation (VQG) dataset~\cite{mostafazadeh2016generating}, which contains human-annotated questions based on images of the MS-COCO dataset. 
This dataset was developed for generating natural and engaging questions based on common sense reasoning. We use the VQG-COCO dataset for our experiments, which contains a total of 2500 training images, 1250 validation images, and 1250 testing images. Each image in the dataset contains five natural questions and five ground truth captions. It is worth noting that the work of~\cite{jain2017creativity} also used the questions from VQA dataset~\cite{VQA} for training purposes, whereas the work by~\cite{mostafazadeh2016generating} uses only the VQG-COCO dataset.
VQA-1.0 dataset is also built on images from the MS-COCO dataset. It contains a total of 82783 images for training, 40504 for validation, and 81434 for testing. Each image is associated with 3 questions. We used a pre-trained caption generation model \cite{Karpathy_CVPR2015} to extract captions for the VQA dataset as the human-annotated captions are not there in the dataset. We train our model separately for VQG-COCO and VQA dataset.

	\begin{algorithm}
	\caption{Multimodal Differential Network}\label{MC-BMN}
	\begin{algorithmic}[1]
		\Procedure{MDN}{$x_{i}$}
		\BState\emph{ Finding Exemplars}:
		\State ${x_i^+,x_i^-}: =KD-Tree(x_{i})$
		\State ${c_i,c_i^+,c_i^- :=} Extract\_caption(x_{i},x_i^+,x_i^-)$
		\BState\emph{ Compute Triplet Embedding}:
		\State  ${g_i,g_i^+,g_i^-}:=Triplet\_CNN(x_{i},x_i^+,x_i^-)$ 
		\State  ${f_i,f_i^+,f_i^-:=}Triplet\_LSTM(c_{i},c_i^+,c_i^-)$

		\BState\emph{ Compute Triplet Fusion Embedding }:
			\State $s_i=Triplet\_Fusion(g_{i},f_{i}, Joint) $
			\State $s^+_i=Triplet\_Fusion(g_{s},f_{s}, Joint)$
			\State $s^-_i=Triplet\_Fusion(g_{c},f_{c}, Joint)$
		\BState\emph{ Compute Triplet Loss}:
		 \State $Loss\_Triplet= triplet\_loss(s_i,s^+_i,s^-_i)$


		\BState \emph{Compute Decode Question Sentence}:
		\State   $\hat{y}=Generating\_LSTM(s_i,h_i,c_i)$
		\State   $loss=Cross\_Entropy(y,\hat{y})$
		\EndProcedure
	\State {-----------------------------------------------------}
	\Procedure{Triplet Fusion}{$g_{i}$,$f_{i},flag$}
			\State $g_{i}$:Image feature,14x14x512
			\State $f_{i}$: Caption feature,1x512
				\BState \emph{Match Dimension}:
			\State $G_{img}=reshape(g_{i})$,196x512
			\State  $F_{caps}=clone(f_{i})$ 196x512
			
			
			\BState \emph{If flag==Joint Fusion}:
            \State $A{jnt}=   \tanh(  W_{ij}  G_{img} \boxdot (W_{cj}  F_{cap} + b_j))$
			\State $S_{emb}=\tanh({W_A} A_{jnt} + b_A)$, 
			\State [$\boxdot = *$  (MDN-Mul), $\boxdot = +$  (MDN-Add)]
			
			\BState \emph{If flag==Attention Fusion }:
			\State $h_{att}= \tanh({W_I}{G_{img}} \eadd ({W_C}{F_{cap}}+{b_c}))$
			\State $P_{att}= \mbox{Softmax}({W_P}{h_{att}}+{b_P})$
			\State $V_{att}= \sum_{i}{P_{att}(i)}{G_{img}(i)}$
			\State $A_{att}= V_{att} + f_{i}$
			\State $S_{emb}=\tanh({W_A} A_{att} + b_A)$
			
			\BState \emph{ Return $S_{emb}$}
			\EndProcedure
	\end{algorithmic}
\end{algorithm}

\subsection{Evaluation Metrics}
Our task is similar to the encoder-decoder framework of machine translation. We have used the same evaluation metric is used in machine translation. BLEU\cite{Papineni_ACL2002} is the first metric to find the correlation between generated questions with ground truth questions. BLEU score is used to measure the precision value, i.e., That is how many words in the predicted question appears in reference question. The BLEU-n score measures the n-gram precision for counting co-occurrence on reference sentences. We have evaluated the BLEU score from n is 1 to 4. The mechanism of ROUGE-n\cite{Lin_ACL2004} score is similar to BLEU-n, whereas, it measures recall value instead of precision value in BLEU. That is how many words in the reference question appears in the predicted question. Another version ROUGE metric is ROUGE-L, which measures the longest common sub-sequence present in the generated question. METEOR\cite{Banerjee_ACL2005} score is another useful evaluation metric to calculate the similarity between generated questions with reference one by considering synonyms, stemming, and paraphrases. The output of the METEOR score measures the word matches between predicted question and reference question. In VQG, it computes the word match score between predicted question with five reference question. CIDer\cite{Vedantam_CVPR2015} score is a consensus-based evaluation metric.  It measures human-likeness, that is the sentence is written by humans or not. The consensus is measured, how often n-grams in the predicted question appeared in the reference question. If the n-grams in the predicted question sentence appear more frequently in reference question, then the question is less informative and have low CIDer score. We provide our results using all these metrics and compare them with existing baselines. 
    
\section{Variations of Proposed Method}
\label{subsec:variants}
While we advocate the use of the multimodal differential network for generating embeddings that can be used by the decoder for generating questions, we also evaluate several variants of this architecture. These are as follows:


\subsection{Differential Image Network}
Instead of using the multimodal differential network for generating embeddings, we also evaluate the differential image network for the same. In this case, the embedding does not include the caption but is based only on the image feature. We also experimented with using multiple exemplars and random exemplars.
For obtaining the exemplar image-based context embedding, we propose a triplet network consist of three networks, one is target net, supporting net, and opposing net.  All these three networks designed with a convolution neural network and shared the same parameters.
\begin{figure}[ht]
	\centering
	\includegraphics[width=0.5\textwidth]{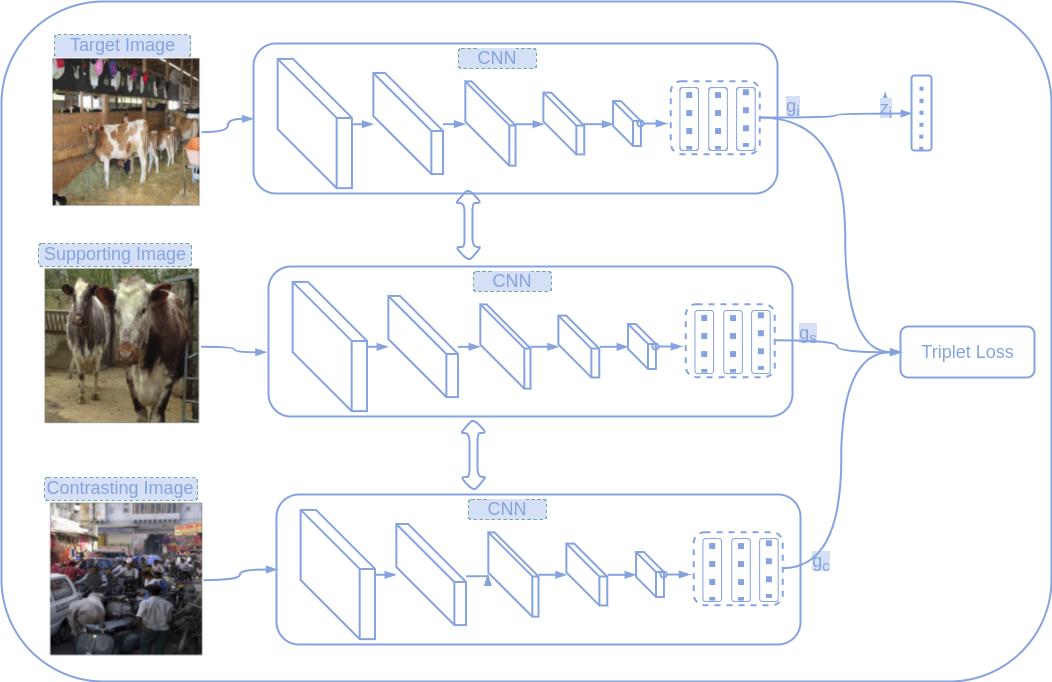}
	\caption{ Differential Image Network }
	\label{fig:DTN}
\end{figure}

The weights of this network are learned through end-to-end learning using a triplet loss.  The aim is to obtain latent weight vectors that bring the supporting exemplar close to the target image and enhances the difference between opposing examples.  More formally, given an image $x_i$, we obtain an embedding $g_i$ using a CNN that we parameterize through a function $G(x_i, W_c)$ where $W_c$ are the weights of the CNN. This is illustrated in  figure~\ref{fig:DTN}.   

\subsection{Tag net}
\label{sec:tag}
\begin{figure}[ht]
	\centering
	\includegraphics[width=0.5\textwidth]{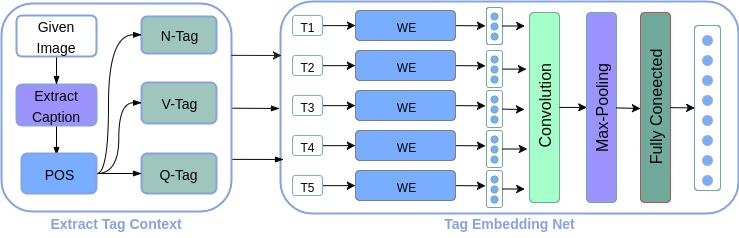}
	\caption{ Illustration of Tag Net }
	\label{fig:tag}
\end{figure}

In this variant, we consider extracting the part-of-speech (POS) tags for the words present in the caption and obtaining a Tag embedding by considering different methods of combining the one-hot vectors.
Basically, it consists of two parts Context Extractor \& Tag Embedding Net. This is illustrated in figure~\ref{fig:tag}.

\textbf{Extract Context}:
The first step is to extract the caption of the image using NeuralTalk2 \cite{Karpathy_NIPS2014} model. We find the part-of-speech(POS) tag present in the caption. POS taggers have been developed for two well-known corpora, the Brown Corpus and the Penn Treebanks. For our work, we are using the Brown Corpus tags. The tags are clustered into three categories, namely Noun tag, Verb tag, and Question tags (What, Where, \dots). Noun tag consists of all the noun \& pronouns present in the caption sentence, and similarly, verb tag consists of verb \& adverbs present in the caption sentence. The question tags consist of the 7-well know question words {i.e., why, how, what, when, where, who, and which}. Each tag token is represented as a one-hot vector of the dimension of vocabulary size. For generalization, we have considered 5 tokens from each category of Tags.

\textbf{Tag Embedding Net}:
The embedding network consists of word embedding followed by temporal convolutions neural network followed by the max-pooling network. In the first step, the sparse high dimensional one-hot vector is transformed into a dense low dimension vector using word embedding. After this, we apply temporal convolution on the word embedding vector. The uni-gram, bi-gram, and tri-gram features are computed by applying a convolution filter of size 1, 2, and 3 respectability. Finally, we applied max-pooling on this to get a vector representation of the tags, as shown figure~\ref{fig:tag}. We concatenated all the tag words followed by a fully connected layer to get a feature dimension of 512. We also explored joint networks based on concatenation of all the tags, on element-wise addition and element-wise multiplication of the tag vectors. However, we observed that convolution over max pooling and joint concatenation gives better performance based on the CIDer score.
\[F_C =\sigma(W_t*\text{T\_CNN}(C_t) + b_t)\]
Where, T\_CNN is Temporally Convolution Neural Network applied on word embedding vector with kernel size three. $\sigma$ is the non-linear activation layer, $W_t, b_t$  are the weight and bias of the corresponding layer.
\subsection{Place net}
In this variant, we explore obtaining embeddings based on the visual scene understanding. This is obtained using a pre-trained PlaceCNN~\cite{Zhou_PAMI2017} that is trained to classify 365 different types of scene categories. We then combine the activation map for the input image and the VGG-19 based place embedding to obtain the joint embedding used by the decoder. 
 Here, places in the image are labeled with scene semantic categories\cite{Zhou_PAMI2017}, comprise of large and diverse types of environment in the world, such as (amusement park, tower, swimming pool, shoe shop, cafeteria, rain-forest, conference center, fish pond, etc.). So we have used a different type of scene semantic categories present in the image as a place-based context to generate a natural question. A place365 is a convolution neural network modeled to classify 365 types of scene categories, which is trained on the place2 dataset consist of 1.8 million scene images. We have used a pre-trained VGG16-places365 network to obtain a place-based context embedding feature for various type scene categories present in the image. The context features $F_C$ is obtained by:
 \[F_C =\sigma(W_p*\text{P\_CNN}(I) + b_p) \]
Where $\text{P\_CNN}$ is Place365\_CNN, $\sigma$ is the non-linear activation layer, $W_p, b_p$  are the weight and bias of the corresponding layer. We have extracted $CONV5$ features of dimension 14x14x512 for attention model and FC8 features of dimension 365 for joint, addition, and Hadamard model of places365. Finally, we use a linear transformation to obtain a 512-dimensional vector. We explored using the CONV$5$, having feature dimension 14x14 512, FC$7$ having 4096, and FC8 having a feature dimension of 365 of places365.

\end{appendices}
\end{document}